\documentclass[sigconf]{acmart}
\AtBeginDocument{%
  }

\setcopyright{acmlicensed}
\copyrightyear{2018}
\acmYear{2018}
\acmDOI{XXXXXXX.XXXXXXX}
\acmConference[Conference acronym 'XX]{Make sure to enter the correct
  conference title from your rights confirmation email}{June 03--05,
  2018}{Woodstock, NY}
  
\usepackage{natbib}
\usepackage{algorithm}
\usepackage{algorithmic}
\usepackage{enumitem}
\usepackage{amsmath}
\usepackage{graphicx} 
\usepackage{array} 
\usepackage{multirow} 
\usepackage{booktabs} 
\usepackage{caption} 
\usepackage{subfig}
\usepackage[dvipsnames]{xcolor}

\usepackage[utf8]{inputenc}
\usepackage{algorithm}
\usepackage{listings}
\usepackage{xcolor}

\definecolor{codeblue}{rgb}{0.25,0.5,0.5}
\definecolor{codekw}{rgb}{0.85, 0.18, 0.50}
\definecolor{codegreen}{rgb}{0,0.6,0}
\definecolor{codegray}{rgb}{0.5,0.5,0.5}
\definecolor{codepurple}{rgb}{0.58,0,0.82}
\definecolor{backcolour}{rgb}{0.95,0.95,0.92}

\lstdefinestyle{mystyle}{
    backgroundcolor=\color{white},   
    commentstyle=\color{codeblue},
    keywordstyle=\color{codekw}\bfseries,
    numberstyle=\tiny\color{codegray},
    stringstyle=\color{codepurple},
    basicstyle=\ttfamily\footnotesize,
    breakatwhitespace=false,         
    breaklines=true,                 
    captionpos=b,                    
    keepspaces=true,                 
    numbers=left,                    
    numbersep=5pt,                  
    showspaces=false,                
    showstringspaces=false,
    showtabs=false,                  
    tabsize=2,
    frame=None
}

\usepackage[table]{xcolor}
\usepackage{multirow}
\usepackage{booktabs}
\usepackage{geometry}
\definecolor{good}{HTML}{E2F0D9} 
\definecolor{great}{HTML}{A9D08E} 
\definecolor{bad}{HTML}{FCE4D6}  
\newcommand{\g}[1]{\cellcolor{good}#1}
\newcommand{\ggc}[1]{\cellcolor{great}#1}
\newcommand{\rbad}[1]{\cellcolor{bad}#1}
\newcommand{\n}{}

\begin{document}

\title{FAiT: Frequency-Aware Inverted Transformer \\ for Multivariate Time Series Forecasting}

\author{Peng He}
\orcid{0009-0008-1664-3963}
\affiliation{%
  \institution{University of Electronic Science and Technology of China}
  \city{Chengdu}
  \state{Sichuan}
  \country{China}}
\email{hepenglk@std.uestc.edu.cn}

\author{Yao Liu}
\affiliation{%
    \institution{University of Electronic Science and Technology of China}
  \city{Chengdu}
  \state{Sichuan}
  \country{China}}
\email{liuyao@uestc.edu.cn}

\author{Yanglei Gan}
\affiliation{%
    \institution{University of Electronic Science and Technology of China}
  \city{Chengdu}
  \state{Sichuan}
  \country{China}}
\email{yangleigan@std.uestc.edu.cn}

\author{Run Lin}
\affiliation{%
    \institution{University of Electronic Science and Technology of China}
  \city{Chengdu}
  \state{Sichuan}
  \country{China}}
\email{runlin@std.uestc.edu.cn}

\author{Yuxiang Cai}
\affiliation{%
    \institution{University of Electronic Science and Technology of China}
  \city{Chengdu}
  \state{Sichuan}
  \country{China}}
\email{yuxiangcai@std.uestc.edu.cn}

\author{Qiao Liu}
\authornote{Corresponding Author.}
\affiliation{%
    \institution{University of Electronic Science and Technology of China}
  \city{Chengdu}
  \state{Sichuan}
  \country{China}}
\email{qliu@uestc.edu.cn}


\begin{abstract}
While Transformer-based architectures have established themselves as a dominant paradigm in Multivariate Time Series Forecasting (MTSF), their core self-attention mechanism inherently functions as a low-pass filter, systematically smoothing out high-frequency signals vital for sharp local changes. Recent advancements have increasingly incorporated frequency-domain operations to address this bias, however, most existing designs rely on fixed spectral bases and apply sequence-wise (uniform) modulation, implicitly assuming a time-invariant frequency response. This overlooks a key property of real-world series that their spectral characteristics often evolve over time, making uniform modulation insufficient for capturing fine-grained temporal dynamics. To tackle these limitations, we propose \textbf{FAiT}, a \textbf{F}requency-\textbf{A}ware \textbf{i}nverted \textbf{T}ransformer. Specifically, FAiT rectifies the spectral bias internally through Inverted Attention, which interprets the attention map as a learnable low-pass operator and constructs a dedicated complementary high-pass branch by inverting the attention matrix to recover attenuated transient signals. Furthermore, FAiT introduces Dynamic Temporal-Frequency Modulation (DTFM), which synthesizes instance-conditioned weights to adaptively re-calibrate the energy of spectral sub-bands, enabling fine-grained control over evolving multi-scale patterns. Extensive experiments on widely used benchmarks demonstrate that FAiT consistently outperforms state-of-the-art Transformer-based and frequency-enhanced baselines, while maintaining computational efficiency. The source code is anonymously online at: \url{https://anonymous.4open.science/r/FAiT-main}.




\end{abstract}


\received{20 February 2007}
\received[revised]{12 March 2009}
\received[accepted]{5 June 2009}

\maketitle

\section{Introduction}

Multivariate time series forecasting (MTSF) transforms raw temporal signals into actionable insights that drive critical decisions across high-stakes domains, from financial risk management \cite{cao2003support,sezer2020financial} and intelligent traffic control \cite{lippi2013short,cirstea2022towards} to climate resilience planning \cite{karevan2020transductive,chen2024marlp} and energy grid optimization \cite{deb2017review,maleki2024future}. In these settings, time series are often high-dimensional, exhibit multi-scale temporal patterns \cite{lim2021time,he2026asynformer} (e.g., trends, seasonality, and abrupt shocks), and display non-stationary behavior \cite{kim2021reversible,liu2023adaptive}. Designing models that can simultaneously capture long-range structure, local fluctuations, and evolving dynamics is therefore technically challenging.

\begin{figure}[t]
	\centering
     \hspace{-0.25cm}
     \subfloat[Self-Attention]{
		\includegraphics[scale=0.262]{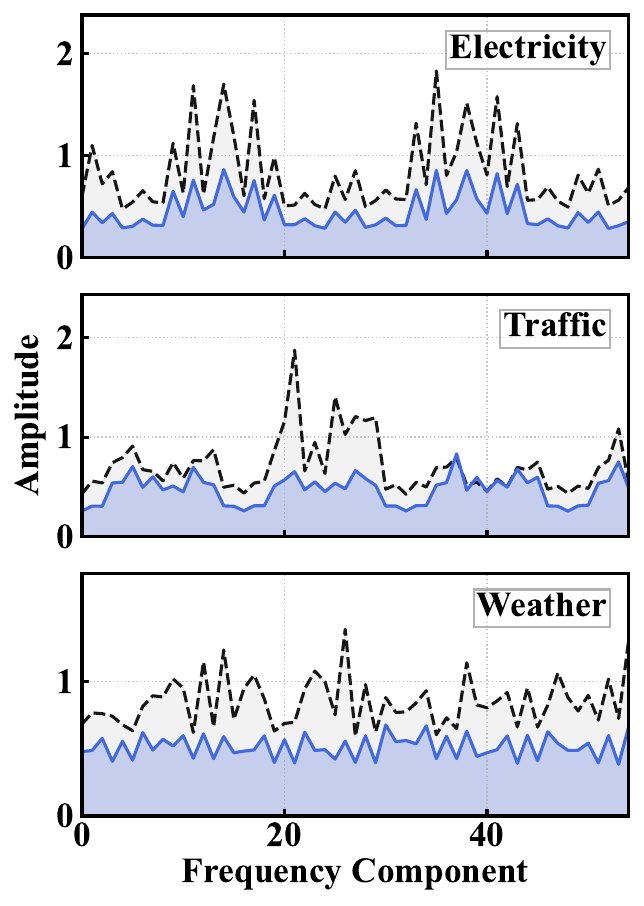}
	}%
  \hspace{-0.25cm}
	   \subfloat[Frequency-Enhanced]{
		\includegraphics[scale=0.262]{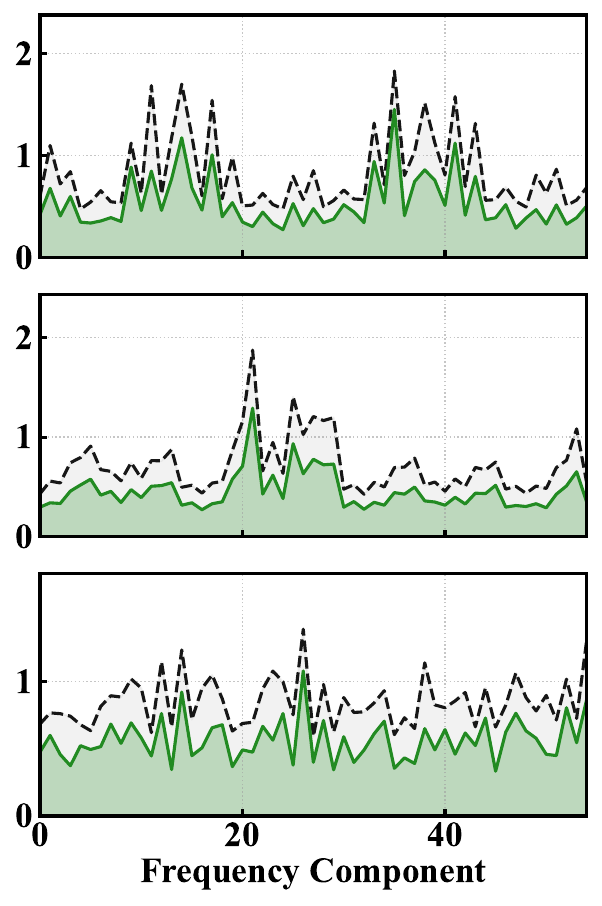}
	}%
  \hspace{-0.25cm}
    \subfloat[Inverted-Attention]{
		\includegraphics[scale=0.262]{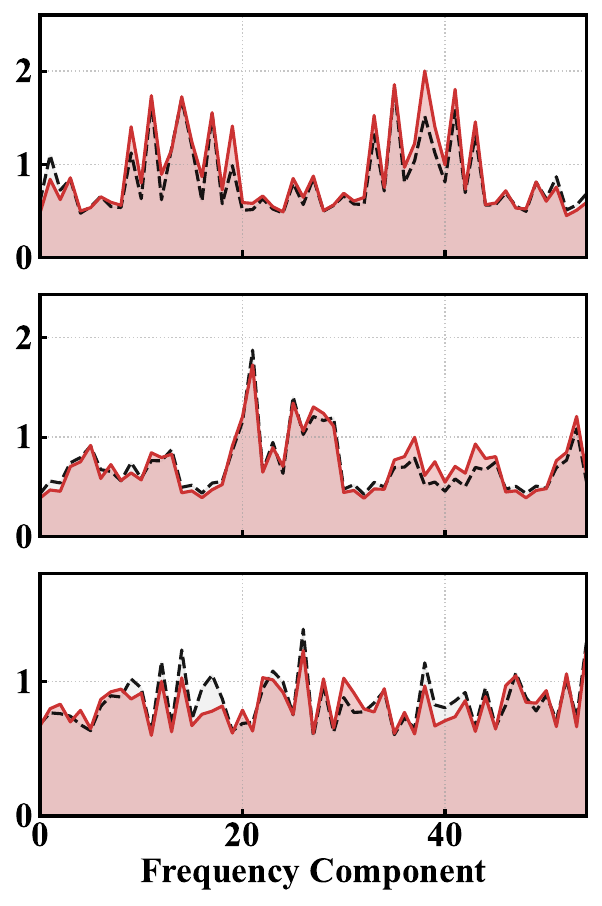}
	}%
    \caption{Spectral energy preservation ratio\protect\footnotemark\ for Self-Attention, Frequency-Enhanced approach (FilterNet) and our Inverted Attention on Electricity, Traffic and Weather datasets. The dashed black curve denotes the ground-truth amplitude spectrum, while the solid colored curve denotes the learned spectrum by models.}
	\label{Visualization_spe}
\end{figure}
\footnotetext{The ratio reports the preserved spectral energy. }


Recent advancements in deep learning, particularly the Transformer architecture \cite{li2019enhancing,zhou2021informer,liu2024itransformer}, have significantly pushed the state-of-the-art in sequence modeling by leveraging the self-attention mechanism to efficiently capture long-range temporal dependencies. However, when applying transformer to time series forecasting tasks, a critical issue emerges. The attention mechanism in transformer exhibits a strong low-pass filtering characteristics, emphasizing slowly varying components while attenuating high-frequency signals \cite{wang22anti,tian2023scan,shin2024attentive,piao2024fredformer,Yue25FreEformer,he2026exploiting}. This bias systematically smooths out sharp local changes and variable-specific high-frequency patterns, which are often crucial for accurate multivariate time series forecasting. Figure~\ref{Visualization_spe} illustrates this phenomenon on three representative datasets (Electricity, Traffic, Weather). The amplitude spectrumof the outputs produced by vanilla self-attention (\textbf{shaded in} \textcolor{blue!70}{\textbf{blue}}) consistently underestimate the peaks of the ground-truth spectrum (\textbf{black dashed}) at medium and high frequencies, confirming the intrinsic low-pass bias. This effect is further amplified by the stacked-layer architecture \cite{zeng2023transformers}, leading to progressive spectral degradation and over-smoothing of the input signal.

To compensate for this, recent research introduce frequency-domain operations into deep models to better exploit the rich spectral structure of time series \cite{yi2023fouriergnn,yi2024filternet,piao2024fredformer,wang2025fredf}. By transforming temporal signals into the frequency domain and manipulating their spectral components, these methods seek to effectively capture trends, seasonal cycles, and other periodic patterns, thereby counteracting the low-pass bias of attention. Despite their empirical success, two critical concerns remain:
\begin{itemize}[leftmargin=*]
    \item \textbf{Fourier-basis projection underfits real-world temporal dynamics.} Most frequency-based methods project the series onto a \textit{fixed basis} \cite{woo2022cost,zhou2022fedformer,yi2024filternet}, constraining all dynamics to a predetermined set of harmonic components. However, real-world signals often exhibit irregular patterns that do not align with these static components \cite{ye2024frequency}. As empirically validated in Figure \ref{Visualization_spe} (b), existing frequency-based methods (\textcolor{ForestGreen!50}{\textbf{green}}) still leave a clear gap in spectral energy preservation relative to the ground truth. This indicates that they largely \textit{\textbf{re-allocate}} predefined harmonics rather than \textit{\textbf{learn}} data-adaptive frequency responses, limiting their ability to match the intrinsic temporal structure.

    \item \textbf{Sequence-wise spectral transforms lack fine-grained temporal modulation.} While several methods explicitly manipulate the \textit{real} and \textit{imaginary} parts of spectral coefficients to retain both amplitude and phase \cite{zhou2022film,xu2024fits,wang2025filterts,Yue25FreEformer}, they typically apply a single spectral operation uniformly to the entire sequence \cite{piao2024fredformer}. This global treatment provides only coarse control. When different time segments emphasize different frequency bands, a fixed sequence-level response cannot modulate frequencies locally in time, limiting its ability to capture fine-grained variations. 

\end{itemize}

To bridge this gap, we propose \textbf{FAiT}, a \textbf{F}requency-\textbf{A}ware \textbf{i}nverted \textbf{T}ransformer that explicitly separates and controls low- and high-frequency behaviors within the attention layers. Specifically, FAiT interprets the standard attention map as a learnable low-pass temporal operator and constructs a complementary high-pass branch by inverting the attention map at each layer. The two branches produce low-frequency and high-frequency feature streams, which are merged through learnable mixing coefficients (Sec. \ref{sec3.2.2}). These coefficients give each layer explicit control over the relative contribution of smooth trends and rapid fluctuations. Figure \ref{Visualization_spe} (c) confirms this design, where Inverted Attention shown in \textcolor{red!50}{\textbf{red}} restores the suppressed medium- and high-frequency components and aligns the model’s spectrum more closely with the target. 

Building on this dual-path design, we introduce dynamic temporal-frequency modulation (DTFM) to provide finer spectral control beyond sequence-wise spectral operations across all time steps. DTFM generates instance-conditioned frequency modulation weights from the fused representation using a lightweight gating network that reassembles a set of learnable spectral prototypes (Sec. \ref{sec3.2.3}). These weights are then applied to the complex spectrum of the hidden states to recalibrate both real and imaginary components before the inverse transform. This design complements attention inversion by improving the modeling of multi-scale patterns and temporally localized variations. Our key contributions are as follows:

\begin{itemize}[leftmargin=*]
    \item We propose FAiT, a Frequency-Aware inverted Transformer for multivariate time series forecasting. FAiT explicitly addresses the low-pass bias of vanilla attention by separating and controlling low- and high-frequency behaviors within attention layers. 
    \item We introduce Dynamic Temporal–Frequency Modulation (DTFM) for fine-grained spectral control. DTFM adaptively reweights temporal–frequency sub-bands based on the current representations, enabling flexible, frequency-dynamic modeling of multi-scale and non-stationary patterns.
    \item We demonstrate that Inverted Attention ($+invAtt$) is a model-agnostic framework. By incorporating this mechanism into diverse Transformer-based forecasters, we show that it serves as a "plug-and-play" booster, consistently unlocking performance gains by restoring high-frequency modeling capabilities.
    \item We extensively compare FAiT with existing methods for multivariate time series forecasting and perform in-depth model analysis. The experiments demonstrate that FAiT consistently excels in a wide range of benchmarks.
\end{itemize}

\section{Related Work}

\subsection{Deep Time Series Forecasting}
Advancements in deep learning have propelled the development of time series forecasting, leading to a surge of innovative models. Among these models, Transformer-based architectures \cite{vaswani2017attention} have garnered particular attention. Early explorations, classic works such as Pyraformer \cite{liu2022pyraformer}, Informer \cite{zhou2021informer} involved initial adaptive upgrades to the original Transformer. Building upon these foundational works, researchers have further optimized the Transformer architecture to better model the specific characteristics of time series data. Representative works include PatchTST's \cite{nie2023a} channel-independent patching, Crossformer's \cite{zhang2023crossformer} two-stage learning across time and variables, CARD's \cite{wang2024card} robust channel alignment strategy, and iTransformer's \cite{liu2024itransformer} inverted embedding attention mechanism. However, the Transformer is not the sole solution. In recent years, Multi-layer perceptron (MLP) architectures, renowned for their lightweight design and efficiency, have also regained widespread academic attention. The genesis of this approach is DLinear \cite{zeng2023transformers}, which revealed the potential of simple linear models to match the performance of complex Transformers, thereby catalyzing a surge in MLP-based research. This led to a series of prominent MLP variants, including TiDE \cite{das2023longterm}, TimeMixer \cite{wang2023timemixer}, TimeMixer++ \cite{wang2025timemixer}, and AMD \cite{hu2025adaptive}. Concurrently, the research paradigms within this field have continued to broaden, with advanced concepts like contrastive learning \cite{poppelbaum2022contrastive,woo2022cost,luo2023time}, causal inference \cite{runge2023causal}, and generative models \cite{yoon2019time,brophy2023generative} being successfully introduced to MTSF tasks.

\begin{figure*}[t]
  \centering
  \includegraphics[width=0.85\linewidth]{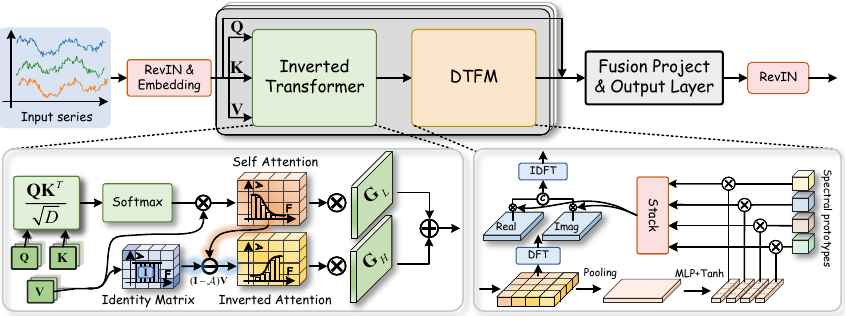}
  \caption{The overall framework of FAiT. The model first employs an \textit{Inverted Transformer} to rectify the inherent smoothing bias of self-attention, utilizing an inverted branch to explicitly recover high-frequency transients suppressed by the low-pass attention map. It then incorporates \textit{Dynamic Temporal-Frequency Modulation (DTFM)}, which synthesizes instance-adaptive weights to dynamically re-calibrate spectral sub-bands, enabling fine-grained modeling of non-stationary patterns.}
  \label{fig:model}
\end{figure*}

\subsection{Frequency Enhanced Paradigm}
Leveraging frequency-domain insights has become an increasingly prominent strategy for enhancing deep learning models in time series forecasting. A growing body of literature demonstrates that employing various frequency-based approaches can lead to superior forecasting capabilities. FEDformer \cite{zhou2022fedformer} adopts a strategy of preprocessing data with a DFT and sampling before feeding it into the Transformer architecture. FiLM \cite{zhou2022film} employs Fourier analysis techniques with the dual objective of preserving historical dependency information in series while effectively mitigating noise interference. FourierGNN \cite{yi2023fouriergnn} transitions the core operations of Graph Neural Networks (GNNs) from the conventional time domain to the frequency domain. FreTS \cite{yi2023frequency} models inter-channel dependencies and temporal correlations in multivariate series by deploying MLP within the frequency domain. Based on FreTS, FilterNet \cite{yi2024filternet} designs learnable filters that selectively pass certain components of the time series signal, thereby extracting critical temporal patterns. FITS \cite{xu2024fits} processes data using a method that combines low-pass filtering with complex-valued linear projection. Fredformer \cite{piao2024fredformer} addresses potential frequency bias by partitioning the frequency spectrum into multiple patches. FAN \cite{ye2024frequency} designs a frequency-adaptive normalization scheme specifically for non-stationary data. FGTI \cite{yang2024frequency} improves the forecasting of primary trend and seasonal components by utilizing high-frequency signals to amplify the residual terms. FreDF \cite{wang2025fredf} introduces a frequency-domain loss function designed to address the label autocorrelation problem that arises in direct multi-step forecasting. FreEformer \cite{Yue25FreEformer} applies the Transformer architecture independently to the real and imaginary components of the frequency spectrum (obtained via DFT) and introduces a learnable parameter matrix to enhance feature diversity. CFPT \cite{kou2025cfpt} constructs a cross-frequency interaction module that enables interplay between different frequency bands, thereby facilitating a modeling approach where long-term trends inform short-term variations. However, most existing methods project dynamics onto fixed basis, which often underfits irregular real-world patterns. Furthermore, they typically apply global spectral operations that lack fine-grained dynamic modulation for localized variations. These limitations motivate our development of a frequency-aware dynamic mechanism, which adaptively calibrates spectral contributions to capture non-stationary dynamics.

\section{Methodology}
\subsection{Preliminaries and Notations}

\textbf{Definition 1 (Multivariate Time Series Forecasting). }The goal of MTSF is to learn a parametric mapping function $\mathrm{F}_{\theta}$ that predicts future observations from historical data. Given an $L$-step history $\mathbf{X}=\{\mathbf{x}_1,\dots,\mathbf{x}_L\} \in \mathbb{R}^{C \times L}$, the model produces an $H$-step forecast $\hat{\mathbf{Y}}=\{\hat{\mathbf{y}}_{L+1},\dots,\hat{\mathbf{y}}_{L+H}\} \in \mathbb{R}^{C \times H}$, where $C$ denotes the number of variables. The parameters $\theta$ are learned by minimizing a loss function $\mathcal{L}$ (e.g. Mean Squared Error (MSE)) that measures the discrepancy between the prediction $\hat{\mathbf{Y}}$ and the ground-truth observations $\mathbf{Y}=\{\mathbf{y}_{L+1},\dots,\mathbf{y}_{L+H}\} \in \mathbb{R}^{C \times H}$:
\begin{equation}
\hat{\mathbf{Y}}= \mathrm{F}_{\theta}(\mathbf{X}),\quad \theta^* = \arg \min_{\theta} \mathcal{L}(\hat{\mathbf{Y}}, \mathbf{Y}).
\end{equation}
\textbf{Definition 2 (Discrete Fourier Transform). }The Discrete Fourier Transform (DFT) is a fundamental tool in digital signal processing \cite{palani2022signals}. Given a length $N$ time-domain sequence $x[n]$, the DFT maps it to the frequency domain via:
\begin{equation}
    \mathcal{X}[k]=\sum_{n=0}^{N-1}x[n]e^{-j2\pi kn/N}, \quad k=0,1,...,N-1,
\end{equation}
where $j$ is the imaginary unit and $\mathcal{X}[k]$ is the complex spectral coefficient associated with the discrete frequency $\omega_k = 2\pi k/N$. Each $\mathcal{X}[k]$ can be decomposed into real and imaginary parts:
\begin{equation}
\begin{aligned} 
    \mathcal{X}[k] = \textbf{Real}(\mathcal{X}[k]) + j\textbf{Imag}(\mathcal{X}[k]), \\
    \textbf{Real}(\mathcal{X}[k])=\sum_{n=0}^{N-1}x[n]\cos\left(\frac{2\pi}{N}kn\right),\\ \textbf{Imag}(\mathcal{X}[k])=-\sum_{n=0}^{N-1}x[n]\sin\left(\frac{2\pi}{N}kn\right). 
\end{aligned}
\end{equation}
The inverse DFT (IDFT) reconstructs the original sequence via:
\begin{equation}
x[n]=\frac{1}{N}\sum_{k=0}^{N-1}\mathcal{X}[k]e^{j2\pi kn/N},\quad n=0,1,\dots,N-1.
\end{equation}
In short, we denote DFT and IDFT operators as $\mathcal{F}$, $\mathcal{F}^{-1}$, respectively.

\subsection{Our Approach}

As aforementioned, existing frequency-based designs can alleviate the low-pass bias of self-attention, but they often rely on fixed basis and global spectral operations, limiting adaptivity to time-varying dynamics. To overcome these limitations, we propose a \textbf{F}requency-\textbf{A}ware \textbf{i}nverted \textbf{T}ransformer (FAiT), which involves two complementary components: (i) an \emph{inverted Transformer} mechanism to explicitly construct low-/high-frequency feature streams within each attention layer; (ii) a \emph{Dynamic Temporal–Frequency Modulation} (DTFM) module to adaptively reweight temporal–frequency sub-bands for fine-grained modeling. An overview of FAiT is provided in Figure \ref{fig:model}, and we detail each component in the following sections.


\subsubsection{\textbf{RevIN and Embedding}}

FAiT first applies Reversible Instance Normalization (RevIN) \cite{kim2021reversible} to normalize the multivariate time series. The corresponding denormalization is performed at the output stage to restore the original scale. After normalization, the input sequence $\mathbf{X} \in \mathbb{R}^{C \times L}$ is projected into a $D$-dimensional embedding space. Concretely, we expand $\mathbf{X}$ by applying a shared linear projection using a learnable matrix $\mathbf{E} \in \mathbb{R}^{1 \times D}$:
\begin{equation}
\mathbf{Z} = \mathbf{X} \cdot \mathbf{E} \in \mathbb{R}^{C \times L \times D} .
\end{equation}

\subsubsection{\textbf{Invert Transformer}}\label{sec3.2.2}
Standard Transformer \cite{vaswani2017attention} architectures rely heavily on the self-attention mechanisms, which aggregates temporal information via weighted averaging. While effective for capturing long-range dependencies, this aggregation creates a significant inductive bias. In this section, we analyze this limitation and introduce \textbf{Inverted Transformer}, designed to decouple and reconstruct the full spectral profile of the time series.

\begin{itemize}[leftmargin=*]

\item\textbf{Theoretical Motivation: The Low-Pass Bias of Attention.}
We first formalize the spectral properties of the self-attention mechanism. Let $\mathbf{Q}$, $\mathbf{K}$, $\mathbf{V}$ represent the query, key, and value projections derived from the input $\mathbf{Z}$. The aggregated representation $\mathbf{Z}_{\text{low}}$ is obtained via the normalized attention matrix $\mathcal{A}$:
\begin{equation}
\mathcal{A} = \mathrm{softmax}\left(\frac{\mathbf{Q}\mathbf{K}^T}{\sqrt{D}}\right), \quad \mathbf{Z}_{\text{low}} = \mathcal{A}\mathbf{V} \in \mathbb{R}^{C \times L \times D}.
\end{equation}
From the perspective of Graph Signal Processing (GSP) \cite{ortega2018graph,choi2024graph,shin2024attentive}, $\mathcal{A}$ functions as a transition matrix. The eigenspectrum of such row-stochastic matrices is dominated by large eigenvalues corresponding to global smoothness (low frequencies), while eigenvalues associated with high-frequency variations decay rapidly \cite{piao2024fredformer}. Consequently, the operation $\mathcal{A}\mathbf{V}$ inherently acts as a low-pass filter, attenuating the high-frequency transients and local abrupt changes that are critical for accurate forecasting.


\item\textbf{High-Frequency Recovery by Inverted Attention.}
To recover the fine-grained spectral details lost during attention aggregation, we introduce a dedicated complementary high-pass branch. Since $\mathcal{A}$ acts as a low-pass filter, its algebraic complement $(\mathbf{I} - \mathcal{A})$ inherently functions as a high-pass filter. We explicitly model the high-frequency component $\mathbf{Z}_{\text{High}}$ as the residual term of the attention operation: 
\begin{equation}
\mathbf{Z}_{\text{High}} = \mathbf{V} - \mathbf{Z}_{\text{Low}} \equiv (\mathbf{I} - \mathcal{A})\mathbf{V} \in \mathbb{R}^{C \times L \times D},
\end{equation}
where $\mathbf{I}$ is the identity matrix. By decomposing the signal into $\mathbf{Z_{Low}}$ and $\mathbf{Z_{High}}$, this method preserves the full signal bandwidth, mitigating the over-smoothing issue typical of Transformers.

\item\textbf{Adaptive Spectral Gating.} In real-world scenarios, the importance of low-frequency trends versus high-frequency details varies dynamically over time. A static summation of $\mathbf{Z_{Low}}$ and $\mathbf{Z_{High}}$ is insufficient for handling non-stationary regimes. We therefore introduce an \textbf{Adaptive Spectral Gating (ASG)} mechanism to dynamically re-calibrate the spectral components. The final output $\tilde{\mathbf{Z}}$ is obtained via a learnable fusion:
\begin{equation}
\begin{gathered}
\mathbf{G}_{L} = \sigma_1(\mathbf{W}_{L} \mathbf{Z} + \mathbf{b}_{L}), \quad \mathbf{G}_{H} = \sigma_2(\mathbf{W}_{H} \mathbf{Z} + \mathbf{b}_{H}),\\
\tilde{\mathbf{Z}} = \mathbf{G}_{L} \odot \mathbf{Z}_{\text{low}} + \mathbf{G}_{H} \odot \mathbf{Z}_{\text{high}} \in \mathbb{R}^{C \times L \times D},
\end{gathered}
\end{equation}
where $\sigma_{1,2}$ are activations, $\odot$ denotes element-wise product, and $\mathbf{W}_{L,H}$ are projections. Algorithm \ref{alg:inverted_attention} provides specific pseudocode. This gating dynamically re-weights spectral contributions per step, harmonizing underlying global trends with local bursts.
\end{itemize}

\subsubsection{\textbf{Dynamic Temporal Frequency Modulation}}\label{sec3.2.3}
While the \textit{Inverted Transformer} module successfully prevents high-frequency attenuation, it operates on a relatively coarse-grained spectral binary decomposition. It only controls the relative contribution of a low-pass component and a complementary high-pass residual. In complex MTSF, however, different variables and even different time segments within the same variable often require nuanced, selective emphasis on specific distinct frequency bands. A straight-forward approach to adjust low/high frequency components is to use static learnable parameters for each band. However, the static approach is sub-optimal given the inherently time-varying and instance-dependent nature of real-world time series. To enable finer and adaptive spectral control, we introduce the \textbf{Dynamic Temporal Frequency Modulation (DTFM)} module, which performs instance-adaptive spectral re-weighting on the fused representation $\tilde{\mathbf{Z}}$. Formally, DTFM first transforms $\tilde{\mathbf{Z}}$ into the frequency domain via $\mathcal{F}(\tilde{\mathbf{Z}})$, applies an instance-conditioned modulation weight, and then reconstructs the refined signal back to the time domain.


\begin{itemize}[leftmargin=*]
    \item \textbf{Frequency Modulation Weights Synthesis.} Instead of learning an independent static scaling parameter for each frequency component, DTFM generates dynamic frequency modulation weights conditioned on $\tilde{\mathbf{Z}}$. Specifically, we define a set of $F$ learnable spectral prototypes $\mathbf{K}=\{\mathbf{K}_1, \dots, \mathbf{K}_F\} \in \mathbb{R}^{F \times (\frac{D}{G}) \times (\frac{L}{2}+1)}$, where each prototype captures a fundamental frequency pattern. The modulation weight is synthesized by reassembling these prototypes according to the input's context. We employ a lightweight MLP with a non-linear activation (Tanh) to learn instance-adaptive coefficients. It maps the learned representation $\tilde{\mathbf{Z}}$ to a dynamic coefficient vector $\alpha$. The modulation weights are constructed as the coefficient-weighted sum of the prototypes:
    \begin{equation}
    \begin{gathered}
    \alpha=\operatorname{Tanh}(\operatorname{MLP}(\operatorname{Avg.Pooling}(\tilde{\mathbf{Z}}))) \in \mathbb{R}^{C \times G \times F}, \\
    \mathcal{W}_{\mathcal{M}} = \operatorname{Reshape}\left( \sum_{f=1}^F \alpha_{:,:,f} \otimes \mathbf{K}_{f} \right) \in \mathbb{R}^{C \times (\frac{L}{2}+1) \times D },
    \end{gathered}
    \end{equation}
    where $\otimes$ denotes the interaction between instance-adaptive coefficients and spectral prototypes, and both $F$ and $G$ are related to the embedding dimension $D$ via the reshaping operation.
    \item \textbf{Dynamic Frequency Modulation.} Utilizing the synthesized real-valued modulation weights $\mathcal{W}_{\mathcal{M}}$, DTFM perform fine-grained spectral modulation in the frequency domain. Let $\mathcal{F}(\tilde{\mathbf{Z}})$ denote the complex spectrum of $\tilde{\mathbf{Z}}$. DTFM applies the same modulation weight to both real and imaginary parts, which preserves the original phase information while adjusting band energy:
\begin{equation}
\begin{aligned}
    \mathbf{Real}'(\tilde{\mathbf{Z}}) &= \mathbf{Real}(\mathcal{F}(\tilde{\mathbf{Z}})) \odot \mathcal{W}_\mathcal{M}(\tilde{\mathbf{Z}}), \\
    \mathbf{Imag}'(\tilde{\mathbf{Z}}) &= \mathbf{Imag}(\mathcal{F}(\tilde{\mathbf{Z}})) \odot \mathcal{W}_\mathcal{M}(\tilde{\mathbf{Z}}).
\end{aligned}
\end{equation}
\end{itemize}
The calibrated spectrum is then reconstructed into the temporal representation $\widehat{\mathbf{Z}}$ via the Inverse transform ($\mathcal{F}^{-1}$):
\begin{equation}
    \widehat{\mathbf{Z}} = \mathcal{F}^{-1}(\mathbf{Real}'(\tilde{\mathbf{Z}}) + j\mathbf{Imag}'(\tilde{\mathbf{Z}})) \in \mathbb{R}^{C \times L \times D}.
\end{equation}


\subsubsection{Residual Fusion and Prediction} Finally, we integrate the spectrally enhanced representation $\widehat{\mathbf{Z}}$ is fused with the module's input $\mathbf{Z}$ via a residual connection to preserve the original feature context. This fused tensor is then processed by the final output stage:
\begin{equation}\begin{gathered} 
    \hat{\mathbf{Y}} = \mathrm{RevIN}^{-1}(\mathrm{Linear}(\mathrm{Flatten}(\widehat{\mathbf{Z}} + \mathbf{Z}))) \in \mathbb{R}^{C \times H}, 
\end{gathered} \end{equation} 
where the linear layer projects the features onto the desired forecast horizon $H$, and $\mathrm{RevIN}^{-1}$ applies the restoration step. 

\begin{table*}[t]
\centering
\fontsize{7.3pt}{7.3pt} \selectfont 
\renewcommand{\arraystretch}{1.2} 
\setlength{\tabcolsep}{1pt} 
\caption{Long-term MTSF performance comparison. We fix the look-back length $L=96$ and report results averaged over horizons $H \in \{96, 192, 336, 720\}$. The best and second-best results are highlighted in \textbf{bold} and \underline{underlined}, respectively.}
\begin{tabular}{@{}c|cccccccccccccccccccccccccc@{}}
\toprule
\multirow{2}{*}{Dataset} & \multicolumn{2}{c}{\begin{tabular}[c]{@{}c@{}}FAiT\\ (Ours)\end{tabular}} & \multicolumn{2}{c}{\begin{tabular}[c]{@{}c@{}}FreDF\\ \citeyearpar{wang2025fredf}\end{tabular}} & \multicolumn{2}{c}{\begin{tabular}[c]{@{}c@{}}FilterTS\\ \citeyearpar{wang2025filterts}\end{tabular}} & \multicolumn{2}{c}{\begin{tabular}[c]{@{}c@{}}FilterNet\\ \citeyearpar{yi2024filternet}\end{tabular}} & \multicolumn{2}{c}{\begin{tabular}[c]{@{}c@{}}FITS\\ \citeyearpar{xu2024fits}\end{tabular}} & \multicolumn{2}{c}{\begin{tabular}[c]{@{}c@{}}CARD\\ \citeyearpar{wang2024card}\end{tabular}} & \multicolumn{2}{c}{\begin{tabular}[c]{@{}c@{}}iTrans. \\ \citeyearpar{liu2024itransformer}\end{tabular}} & \multicolumn{2}{c}{\begin{tabular}[c]{@{}c@{}}TimeMixer\\ \citeyearpar{wang2023timemixer}\end{tabular}} & \multicolumn{2}{c}{\begin{tabular}[c]{@{}c@{}}PatchTST\\ \citeyearpar{nie2023a}\end{tabular}} & \multicolumn{2}{c}{\begin{tabular}[c]{@{}c@{}}Crossfm.\\ \citeyearpar{zhang2023crossformer}\end{tabular}} & \multicolumn{2}{c}{\begin{tabular}[c]{@{}c@{}}TimesNet\\ \citeyearpar{wu2023timesnet}\end{tabular}} & \multicolumn{2}{c}{\begin{tabular}[c]{@{}c@{}}FreTS\\ \citeyearpar{yi2023frequency}\end{tabular}} & \multicolumn{2}{c}{\begin{tabular}[c]{@{}c@{}}DLinear\\ \citeyearpar{zeng2023transformers}\end{tabular}} \\ \cmidrule(l){2-27}
 & MSE & \multicolumn{1}{c|}{MAE} & MSE & \multicolumn{1}{c|}{MAE} & MSE & \multicolumn{1}{c|}{MAE} & MSE & \multicolumn{1}{c|}{MAE} & MSE & \multicolumn{1}{c|}{MAE} & MSE & \multicolumn{1}{c|}{MAE} & MSE & \multicolumn{1}{c|}{MAE} & MSE & \multicolumn{1}{c|}{MAE} & MSE & \multicolumn{1}{c|}{MAE} & MSE & \multicolumn{1}{c|}{MAE} & MSE & \multicolumn{1}{c|}{MAE} & MSE & \multicolumn{1}{c|}{MAE} & MSE & MAE \\ \midrule
ETTm1 & \textbf{0.378} & \multicolumn{1}{c|}{\textbf{0.380}} & 0.392 & \multicolumn{1}{c|}{0.399} & 0.385 & \multicolumn{1}{c|}{0.396} & 0.392 & \multicolumn{1}{c|}{0.401} & 0.493 & \multicolumn{1}{c|}{0.452} & 0.383 & \multicolumn{1}{c|}{\underline{ 0.384}} & 0.407 & \multicolumn{1}{c|}{0.410} & \underline{ 0.381} & \multicolumn{1}{c|}{0.395} & 0.387 & \multicolumn{1}{c|}{0.400} & 0.513 & \multicolumn{1}{c|}{0.496} & 0.400 & \multicolumn{1}{c|}{0.406} & 0.407 & \multicolumn{1}{c|}{0.415} & 0.403 & 0.407 \\
ETTm2 & \textbf{0.271} & \multicolumn{1}{c|}{\textbf{0.314}} & 0.278 & \multicolumn{1}{c|}{0.319} & 0.276 & \multicolumn{1}{c|}{0.321} & 0.285 & \multicolumn{1}{c|}{0.328} & 0.291 & \multicolumn{1}{c|}{0.333} & \underline{ 0.272} & \multicolumn{1}{c|}{\underline{ 0.317}} & 0.288 & \multicolumn{1}{c|}{0.332} & 0.275 & \multicolumn{1}{c|}{0.323} & 0.281 & \multicolumn{1}{c|}{0.326} & 0.757 & \multicolumn{1}{c|}{0.610} & 0.291 & \multicolumn{1}{c|}{0.333} & 0.335 & \multicolumn{1}{c|}{0.379} & 0.350 & 0.401 \\
ETTh1 & \textbf{0.430} & \multicolumn{1}{c|}{\textbf{0.426}} & 0.437 & \multicolumn{1}{c|}{0.435} & \underline{ 0.433} & \multicolumn{1}{c|}{0.430} & 0.441 & \multicolumn{1}{c|}{0.439} & 0.440 & \multicolumn{1}{c|}{0.431} & 0.442 & \multicolumn{1}{c|}{\underline{ 0.429}} & 0.454 & \multicolumn{1}{c|}{0.447} & 0.447 & \multicolumn{1}{c|}{0.440} & 0.469 & \multicolumn{1}{c|}{0.454} & 0.529 & \multicolumn{1}{c|}{0.522} & 0.458 & \multicolumn{1}{c|}{0.450} & 0.488 & \multicolumn{1}{c|}{0.474} & 0.456 & 0.452 \\
ETTh2 & \underline{ 0.365} & \multicolumn{1}{c|}{\textbf{0.389}} & 0.371 & \multicolumn{1}{c|}{0.396} & 0.372 & \multicolumn{1}{c|}{0.396} & 0.383 & \multicolumn{1}{c|}{0.407} & 0.376 & \multicolumn{1}{c|}{0.398} &  0.368 & \multicolumn{1}{c|}{\underline{ 0.390}} & 0.383 & \multicolumn{1}{c|}{0.407} & \textbf{0.364} & \multicolumn{1}{c|}{0.395} & 0.384 & \multicolumn{1}{c|}{0.405} & 0.942 & \multicolumn{1}{c|}{0.684} & 0.414 & \multicolumn{1}{c|}{0.427} & 0.550 & \multicolumn{1}{c|}{0.515} & 0.559 & 0.515 \\
ECL & \textbf{0.164} & \multicolumn{1}{c|}{\textbf{0.252}} & 0.170 & \multicolumn{1}{c|}{0.259} & 0.180 & \multicolumn{1}{c|}{0.271} & 0.173 & \multicolumn{1}{c|}{0.268} & 0.384 & \multicolumn{1}{c|}{0.434} & \underline{ 0.168} & \multicolumn{1}{c|}{\underline{ 0.258}} & 0.178 & \multicolumn{1}{c|}{0.270} & 0.182 & \multicolumn{1}{c|}{0.272} & 0.208 & \multicolumn{1}{c|}{0.295} & 0.244 & \multicolumn{1}{c|}{0.334} & 0.192 & \multicolumn{1}{c|}{0.295} & 0.202 & \multicolumn{1}{c|}{0.290} & 0.212 & 0.300 \\
Exchange & \textbf{0.342} & \multicolumn{1}{c|}{\textbf{0.393}} & 0.387 & \multicolumn{1}{c|}{0.421} & \underline{ 0.352} & \multicolumn{1}{c|}{\underline{ 0.397}} & 0.388 & \multicolumn{1}{c|}{0.419} & 0.365 & \multicolumn{1}{c|}{0.408} & 0.362 & \multicolumn{1}{c|}{0.402} & 0.360 & \multicolumn{1}{c|}{0.403} & 0.387 & \multicolumn{1}{c|}{0.416} & 0.367 & \multicolumn{1}{c|}{0.404} & 0.940 & \multicolumn{1}{c|}{0.707} & 0.416 & \multicolumn{1}{c|}{0.443} & 0.416 & \multicolumn{1}{c|}{0.439} & 0.354 & 0.414 \\
Traffic & 0.436 & \multicolumn{1}{c|}{\textbf{0.258}} & \textbf{0.421} & \multicolumn{1}{c|}{\underline{ 0.279}} & 0.471 & \multicolumn{1}{c|}{0.315} & 0.463 & \multicolumn{1}{c|}{0.310} & 0.615 & \multicolumn{1}{c|}{0.370} & 0.453 & \multicolumn{1}{c|}{0.282} & \underline{ 0.428} & \multicolumn{1}{c|}{0.282} & 0.484 & \multicolumn{1}{c|}{0.297} & 0.531 & \multicolumn{1}{c|}{0.343} & 0.550 & \multicolumn{1}{c|}{0.304} & 0.620 & \multicolumn{1}{c|}{0.336} & 0.538 & \multicolumn{1}{c|}{0.328} & 0.625 & 0.383 \\
Weather & \textbf{0.239} & \multicolumn{1}{c|}{\textbf{0.261}} & 0.254 & \multicolumn{1}{c|}{0.274} & 0.244 & \multicolumn{1}{c|}{0.274} & 0.245 & \multicolumn{1}{c|}{0.272} & 0.273 & \multicolumn{1}{c|}{0.292} & \textbf{0.239} & \multicolumn{1}{c|}{\underline{ 0.265}} & 0.258 & \multicolumn{1}{c|}{0.279} & \underline{ 0.240} & \multicolumn{1}{c|}{0.271} & 0.259 & \multicolumn{1}{c|}{0.281} & 0.259 & \multicolumn{1}{c|}{0.315} & 0.259 & \multicolumn{1}{c|}{0.287} & 0.255 & \multicolumn{1}{c|}{0.298} & 0.265 & 0.317 \\ \midrule
$1^{st}$ Count & \textbf{6} & \multicolumn{1}{c|}{\textbf{8}} & 1 & \multicolumn{1}{c|}{0} & 0 & \multicolumn{1}{c|}{0} & 0 & \multicolumn{1}{c|}{0} & 0 & \multicolumn{1}{c|}{0} & 1 & \multicolumn{1}{c|}{0} & 0 & \multicolumn{1}{c|}{0} & 1 & \multicolumn{1}{c|}{0} & 0 & \multicolumn{1}{c|}{0} & 0 & \multicolumn{1}{c|}{0} & 0 & \multicolumn{1}{c|}{0} & 0 & \multicolumn{1}{c|}{0} & 0 & 0 \\ \bottomrule
\end{tabular}
\label{over_long_per}
\end{table*}

\begin{table*}[h!]
\centering
\fontsize{7.3pt}{8.6pt} \selectfont 
\renewcommand{\arraystretch}{1} 
\setlength{\tabcolsep}{2.7pt} 
\caption{Short-term MTSF performance comparison. With look-back length $L=96$, results are averaged over horizons $H \in \{12, 24, 48, 96\}$. Other details are consistent with Table \ref{over_long_per}.}
\begin{tabular}{@{}c|cc|cc|cc|cc|cc|cc|cc|cc|cc|cc|cc@{}}
\toprule
\multirow{2}{*}{Dataset} & \multicolumn{2}{c}{\begin{tabular}[c]{@{}c@{}}FAiT\\ (Ours)\end{tabular}} & \multicolumn{2}{c}{\begin{tabular}[c]{@{}c@{}}TimeMixer++\\ \citeyearpar{wang2025timemixer}\end{tabular}} & \multicolumn{2}{c}{\begin{tabular}[c]{@{}c@{}}FilterNet\\ \citeyearpar{yi2024filternet}\end{tabular}} & \multicolumn{2}{c}{\begin{tabular}[c]{@{}c@{}}FITS\\ \citeyearpar{xu2024fits}\end{tabular}} & \multicolumn{2}{c}{\begin{tabular}[c]{@{}c@{}}CARD\\ \citeyearpar{wang2024card}\end{tabular}} & \multicolumn{2}{c}{\begin{tabular}[c]{@{}c@{}}Fredfm.\\ \citeyearpar{piao2024fredformer}\end{tabular}} & \multicolumn{2}{c}{\begin{tabular}[c]{@{}c@{}}TimeMixer\\ \citeyearpar{wang2023timemixer}\end{tabular}} & \multicolumn{2}{c}{\begin{tabular}[c]{@{}c@{}}PatchTST\\ \citeyearpar{nie2023a}\end{tabular}} & \multicolumn{2}{c}{\begin{tabular}[c]{@{}c@{}}TimesNet\\ \citeyearpar{wu2023timesnet}\end{tabular}} & \multicolumn{2}{c}{\begin{tabular}[c]{@{}c@{}}Crossfm.\\ \citeyearpar{zhang2023crossformer}\end{tabular}} & \multicolumn{2}{c}{\begin{tabular}[c]{@{}c@{}}Dlinear\\ \citeyearpar{zeng2023transformers}\end{tabular}} \\ \cmidrule(l){2-23} 
 & MSE & MAE & MSE & MAE & MSE & MAE & MSE & MAE & MSE & MAE & MSE & MAE & MSE & MAE & MSE & MAE & MSE & MAE & MSE & MAE & MSE & MAE \\ \midrule
PEMS03 & \textbf{0.131} & \textbf{0.233} & 0.165 & 0.263 & 0.145 & 0.251 & 0.489 & 0.465 & 0.174 & 0.275 & \underline{0.135} & \underline{ 0.243} & 0.167 & 0.267 & 0.180 & 0.291 & 0.147 & 0.248 & 0.169 & 0.282 & 0.278 & 0.375 \\
PEMS04 & \textbf{0.120} & \textbf{0.225} & 0.136 & 0.251 & 0.146 & 0.258 & 0.531 & 0.489 & 0.206 & 0.299 & 0.162 & 0.261 & 0.185 & 0.287 & 0.195 & 0.307 & \underline{ 0.129} & \underline{ 0.241} & 0.209 & 0.314 & 0.295 & 0.388 \\
PEMS07 & \textbf{0.097} & \textbf{0.191} & 0.152 & 0.258 & 0.123 & 0.229 & 0.500 & 0.472 & 0.149 & 0.247 & \underline{ 0.121} & \underline{ 0.222} & 0.181 & 0.271 & 0.211 & 0.303 & 0.124 & 0.225 & 0.235 & 0.315 & 0.329 & 0.396 \\
PEMS08 & \textbf{0.146} & \textbf{0.229} & 0.200 & 0.279 & 0.172 & 0.260 & 0.534 & 0.487 & 0.201 & 0.280 & \underline{ 0.161} & \underline{ 0.250} & 0.226 & 0.299 & 0.280 & 0.321 & 0.193 & 0.271 & 0.268 & 0.307 & 0.379 & 0.416 \\ \bottomrule
\end{tabular}
\label{over_short_pre}
\end{table*}

\section{Experiments}
\subsection{Experimental Setups}
\textbf{Datasets. } We evaluate FAiT on 12 widely-used multivariate time series datasets, encompassing both long-term and short-term forecasting scenarios. These include: (i) \textbf{Long-term benchmarks:} ETT (m1, m2, h1, h2), Electricity (ECL), Exchange, Traffic, and Weather; and (ii) \textbf{Short-term benchmarks:} PEMS (03, 04, 07, 08). 


\noindent\textbf{Baselines. } We benchmark FAiT against 14 state-of-the-art methods across three distinct categories:
(i) \textbf{Linear-based models:} TimeMixer++ \cite{wang2025timemixer}, TimeMixer \cite{wang2023timemixer}, TimesNet \cite{wu2023timesnet}, DLinear \cite{zeng2023transformers};
(ii) \textbf{Transformer-based models:} CARD \cite{wang2024card}, iTransformer \cite{liu2024itransformer}, PatchTST \cite{nie2023a}, Crossformer \cite{zhang2023crossformer};
(iii) \textbf{Frequency-domain models:} FreDF \cite{wang2025fredf}, FilterTS \cite{wang2025filterts}, FilterNet \cite{yi2024filternet}, FITS \cite{xu2024fits}, Fredformer \cite{piao2024fredformer}, FreTS \cite{yi2023frequency}. 


\noindent\textbf{Evaluation Metrics. }Performance is assessed using five standard metrics: Mean Absolute Error (MAE), Mean Squared Error (MSE), Coefficient of Determination ($R^2$), the Pearson Correlation Coefficient ($r$), and Mean Absolute Scaled Error (MASE). 


\noindent\textbf{Implementation Details. }All experiments are implemented in PyTorch and conducted on a single NVIDIA A100 GPU (80 GB). To ensure fair comparison, we align our experimental protocol with CARD \cite{wang2024card}, utilizing the L1 loss function. The model is optimized via Adam, with hyperparameters determined through grid search. Learning rates are selected from $\{\text{5e-4}, \text{2e-4},\text{1e-4}\}$ and the embedding dimension $D$ from $\{\text{8}, \text{16},\text{32},\text{64},\text{128}\}$. 


\begin{table*}[t]
\centering
\fontsize{9pt}{6pt} \selectfont 
\renewcommand{\arraystretch}{0.7} 
\setlength{\tabcolsep}{3pt} 
\caption{Forecasting results obtained by applying our proposed Inverted Attention ($+invAtt$) to four mainstream Transformer-based baselines under four prediction horizons $H=\{96,192,336,720\}$. The look-back window $L$ is fixed at 336. The highlighted cells indicate improvements brought by $invAtt$, where darker shades correspond to larger performance gains.}
\begin{tabular}{@{}cc|cccc|cccc|cccc|cccc@{}}
\toprule
\multicolumn{1}{c|}{\multirow{2}{*}{Dataset}}     & Model        & \multicolumn{2}{c}{CARD} & \multicolumn{2}{c|}{$+invAtt$} & \multicolumn{2}{c}{iTrans.} & \multicolumn{2}{c|}{$+invAtt$} & \multicolumn{2}{c}{PatchTST} & \multicolumn{2}{c|}{$+invAtt$} & \multicolumn{2}{c}{Trans.} & \multicolumn{2}{c}{$+invAtt$} \\ \cmidrule(l){2-18} 
\multicolumn{1}{c|}{}                             & Metrics      & MSE         & MAE        & MSE              & MAE              & MSE             & MAE            & MSE                  & MAE                  & MSE           & MAE          & MSE                & MAE                & MSE            & MAE            & MSE                 & MAE                 \\ \midrule
\multicolumn{1}{c|}{\multirow{5}{*}{ETTh1}}       & 96           & 0.369       & 0.393      & \n 0.369         & \g 0.388         & 0.402           & 0.418          & \g 0.396             & \g 0.414             & 0.385         & 0.408        & \g 0.378           & \g 0.401           & 1.038          & 0.826          & \g 1.002            & \rbad 0.835         \\
\multicolumn{1}{c|}{}                             & 192          & 0.411       & 0.421      & \g 0.408         & \g 0.411         & 0.450           & 0.449          & \g 0.437             & \g 0.441             & 0.485         & 0.475        & \ggc 0.446         & \ggc 0.441         & 1.378          & 0.974          & \ggc 1.105          & \ggc 0.821          \\
\multicolumn{1}{c|}{}                             & 336          & 0.438       & 0.437      & \g 0.432         & \g 0.429         & 0.479           & 0.470          & \ggc 0.456           & \g 0.458             & 0.512         & 0.494        & \ggc 0.462         & \ggc 0.458         & 1.213          & 0.884          & \ggc 1.101          & \ggc 0.831          \\
\multicolumn{1}{c|}{}                             & 720          & 0.451       & 0.469      & \g 0.434         & \g 0.450         & 0.584           & 0.548          & \ggc 0.502           & \ggc 0.504           & 0.556         & 0.527        & \ggc 0.476         & \ggc 0.492         & 1.239          & 0.921          & \ggc 1.112          & \ggc 0.851          \\ \cmidrule(l){2-18} 
\multicolumn{1}{c|}{}                             & \textbf{Avg.} & 0.417       & 0.430      & \g 0.411         & \g 0.420         & 0.479           & 0.471          & \ggc 0.448           & \g 0.454             & 0.485         & 0.476        & \ggc 0.441         & \ggc 0.448         & 1.217          & 0.901          & \ggc 1.080          & \ggc 0.835          \\ \midrule
\multicolumn{1}{c|}{\multirow{5}{*}{ETTm1}}       & 96           & 0.291       & 0.338      & \g 0.286         & \g 0.329         & 0.303           & 0.357          & \g 0.302             & \g 0.355             & 0.295         & 0.345        & \g 0.291           & \n 0.347           & 0.871          & 0.696          & \ggc 0.665          & \ggc 0.607          \\
\multicolumn{1}{c|}{}                             & 192          & 0.329       & 0.364      & \rbad 0.331      & \g 0.356         & 0.345           & 0.383          & \g 0.342             & \g 0.380             & 0.337         & 0.379        & \g 0.331           & \g 0.372           & 0.734          & 0.643          & \g 0.721            & \g 0.637            \\
\multicolumn{1}{c|}{}                             & 336          & 0.370       & 0.387      & \g 0.367         & \g 0.376         & 0.382           & 0.405          & \g 0.377             & \g 0.400             & 0.367         & 0.393        & \g 0.364           & \g 0.391           & 1.031          & 0.806          & \ggc 0.836          & \ggc 0.716          \\
\multicolumn{1}{c|}{}                             & 720          & 0.430       & 0.422      & \g 0.421         & \g 0.412         & 0.443           & 0.439          & \g 0.439             & \g 0.437             & 0.428         & 0.435        & \g 0.421           & \g 0.421           & 0.935          & 0.806          & \g 0.921            & \g 0.789            \\ \cmidrule(l){2-18} 
\multicolumn{1}{c|}{}                             & \textbf{Avg.} & 0.355       & 0.378      & \g 0.351         & \g 0.368         & 0.368           & 0.396          & \g 0.365             & \g 0.393             & 0.357         & 0.388        & \g 0.352           & \g 0.383           & 0.893          & 0.738          & \ggc 0.786          & \ggc 0.687          \\ \midrule
\multicolumn{1}{c|}{\multirow{5}{*}{Electricity}} & 96           & 0.128       & 0.222      & 0.128            & \g 0.221         & 0.133           & 0.229          & \g 0.131             & \g 0.226             & 0.138         & 0.241        & \g 0.134           & \g 0.239           & 0.353          & 0.441          & \ggc 0.314          & \ggc 0.415          \\
\multicolumn{1}{c|}{}                             & 192          & 0.153       & 0.246      & \ggc 0.138       & \g 0.235         & 0.156           & 0.251          & \g 0.153             & \g 0.247             & 0.154         & 0.255        & \g 0.152           & \g 0.249           & 0.382          & 0.493          & \g 0.375            & \ggc 0.421          \\
\multicolumn{1}{c|}{}                             & 336          & 0.168       & 0.262      & \g 0.162         & \g 0.257         & 0.172           & 0.267          & \g 0.170             & \g 0.265             & 0.171         & 0.273        & \g 0.163           & \g 0.265           & 0.522          & 0.546          & \g 0.516            & \g 0.538            \\
\multicolumn{1}{c|}{}                             & 720          & 0.207       & 0.292      & \g 0.197         & \g 0.287         & 0.209           & 0.304          & \g 0.205             & \g 0.301             & 0.212         & 0.303        & \g 0.202           & \g 0.297           & 0.477          & 0.519          & \ggc 0.438          & \ggc 0.493          \\ \cmidrule(l){2-18} 
\multicolumn{1}{c|}{}                             & \textbf{Avg.} & 0.164       & 0.256      & \g 0.156         & \g 0.250         & 0.168           & 0.263          & \g 0.165             & \g 0.260             & 0.169         & 0.268        & \g 0.163           & \g 0.263           & 0.434          & 0.500          & \ggc 0.411          & \ggc 0.467          \\ \midrule
\multicolumn{2}{c|}{\textbf{AVG Improve (\%)}}                   & \textbf{-}  & \textbf{-} & \textbf{2.50\%}  & \textbf{2.37\%}  & \textbf{-}      & \textbf{-}     & \textbf{3.00\%}      & \textbf{1.84\%}      & \textbf{-}    & \textbf{-}   & \textbf{4.68\%}    & \textbf{3.10\%}    & \textbf{-}     & \textbf{-}     & \textbf{9.50\%}     & \textbf{6.95\%}     \\ \bottomrule
\end{tabular}
\label{eff_invertatt}
\end{table*}

\subsection{Main Results}

\paragraph{\textbf{Long-Term Time Series Forecasting}} We present the averaged performance of FAiT under four forecasting horizons compared to mainstream baselines in Table \ref{over_long_per}. FAiT consistently produces highly accurate forecasts across all eight datasets, achieving best or second-best performance in 27 (19/8) out of 32 cases for MSE and 31 (26/5) out of 32 cases for MAE, which evidences its stability across diverse domains. Notably, these gains hold against recent frequency-enhanced models such as FreDF, FilterTS, FilterNet, indicating that FAiT provides a more effective way to exploit frequency-related information than existing spectral or filtering-based designs. At the dataset level, FAiT establishes a new state of the art on six out of eight datasets in terms of MSE and on all eight datasets in terms of MAE, underscoring its strong advantage for long-term time series forecasting. 

\paragraph{\textbf{Short-Term Time Series Forecasting}} Table \ref{over_short_pre} summarizes the short-term forecasting results on four PEMS traffic datasets. FAiT achieves the lowest MSE and MAE on all four datasets, showing consistent superiority over both transformer-based and frequency-enhanced baselines. Compared with the strongest competitor on each dataset (typically FreDfm or TimesNet), FAiT reduces MSE by about 3\%, 7\%, 20\% and 9\% on PEMS03/04/07/08, respectively, and lowers MAE by roughly 4\%, 7\%, 14\% and 8\%. The largest gains appear on PEMS07, which has the widest spatial coverage (nearly 900 sensors) and thus the most complex traffic patterns, indicating that FAiT scales well to high-dimensional settings and remains effective under highly volatile urban traffic dynamics.

\paragraph{\textbf{Framework Generality}} We apply our proposed Inverted Attention ($+invAtt$) to four mainstream Transformer-based methods and report the performance improvement of each model (Table \ref{eff_invertatt}). Our module consistently enhances the forecasting ability of different backbones. Overall, it achieves averaged 9.50\% and 6.95\% performance gains on Transformer on MSE and MAE respectively. Beyond the vanilla Transformer, CARD, iTransformer and PatchTST also benefit from the plug-and-play design, obtaining average relative improvements of 2.50\%/2.37\%, 3.00\%/1.84\% and 4.68\%/3.10\% on MSE/MAE, respectively. Notably, gains are more pronounced on the hourly-sampled ETTh1 than on the 15-minute-sampled ETTm1. Although the input length is fixed at 336 for both datasets, 336 hourly steps in ETTh1 cover a much longer temporal span than 336 15-minute steps in ETTm1, making long-range daily and multi-day dependencies more dominant and rendering high-frequency variations weaker and easier to over-smooth for standard Transformers with an intrinsic low-pass bias. By explicitly reinforcing these under-represented local fluctuations, our Inverted Attention mechanism yields substantial additional gains on ETTh1, effectively compensating for the loss of fine-grained temporal detail that conventional architectures suffer at coarser temporal resolutions. Extended evaluations on additional metrics ($R^2$, MASE, $r$) further confirm FAiT's superiority and strong generalization across most scenarios.



\begin{table}[t]
\centering
\fontsize{7pt}{7pt} \selectfont 
\renewcommand{\arraystretch}{1.2} 
\setlength{\tabcolsep}{1pt} 
\caption{Average ablation results with $L = 96$ across the forecast horizons $H \in \{96, 192, 336, 720\}$ on six datasets.}
\begin{tabular}{@{}c|cccccccccccc@{}}
\toprule
Dataset & \multicolumn{2}{c}{ETTm1} & \multicolumn{2}{c}{ETTm2} & \multicolumn{2}{c}{ETTh1} & \multicolumn{2}{c}{ETTh2} & \multicolumn{2}{c}{ECL} & \multicolumn{2}{c}{Exchange} \\ \midrule
Metrics & MSE & MAE & MSE & MAE & MSE & MAE & MSE & MAE & MSE & MAE & MSE & MAE \\ \midrule
w/o DTFM & 0.387 & 0.386 & 0.275 & 0.318 & 0.435 & 0.430 & 0.369 & 0.392 & 0.173 & 0.258 & 0.365 & 0.404 \\
w/o invAtt & 0.388 & 0.388 & 0.279 & 0.317 & 0.432 & 0.428 & 0.367 & 0.391 & 0.173 & 0.258 & 0.360 & 0.404 \\
Re $\mathbf{Z}_{\text{high}}$ & 0.387 & 0.387 & 0.279 & 0.317 & 0.432 & 0.428 & 0.367 & 0.391 & 0.172 & 0.257 & 0.362 & 0.402 \\ \midrule
w/o $\mathcal{W}_\mathcal{M}$ & 0.391 & 0.387 & 0.278 & 0.317 & 0.435 & 0.430 & 0.368 & 0.391 & 0.172 & 0.258 & 0.358 & 0.402 \\
w/o ASG & 0.388 & 0.387 & 0.278 & 0.317 & 0.432 & 0.428 & 0.368 & 0.392 & 0.174 & 0.259 & 0.355 & 0.399 \\ \midrule
FAiT & 0.378 & 0.380 & 0.271 & 0.314 & 0.431 & 0.426 & 0.365 & 0.389 & 0.164 & 0.252 & 0.342 & 0.393 \\ \bottomrule
\end{tabular}
\label{abla}
\end{table}

\begin{figure}[]
	\centering
   \subfloat[FAiT (96 $\to$ 192)]{
		\includegraphics[scale=0.16]{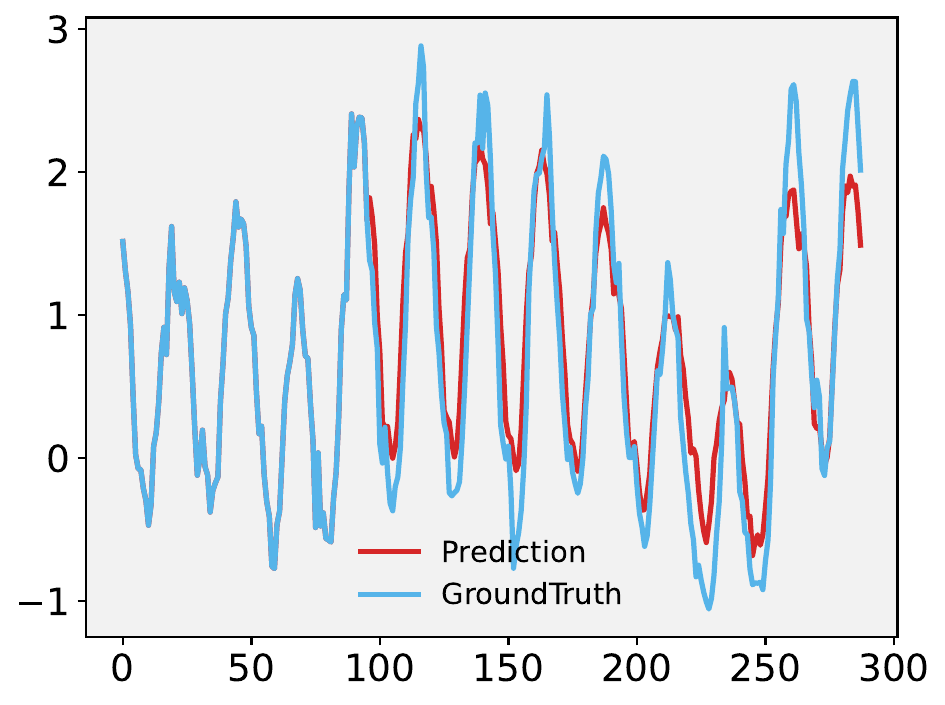}
	}%
 	\subfloat[w/o DTFM]{
		\includegraphics[scale=0.16]{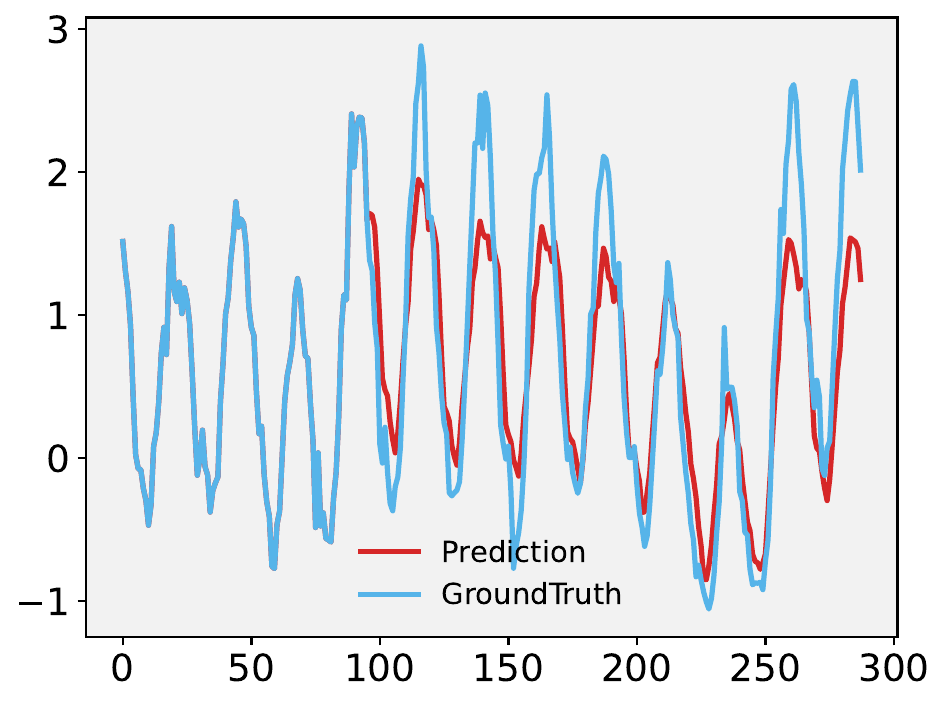}
	}%
 	\subfloat[w/o invAtt]{
		\includegraphics[scale=0.16]{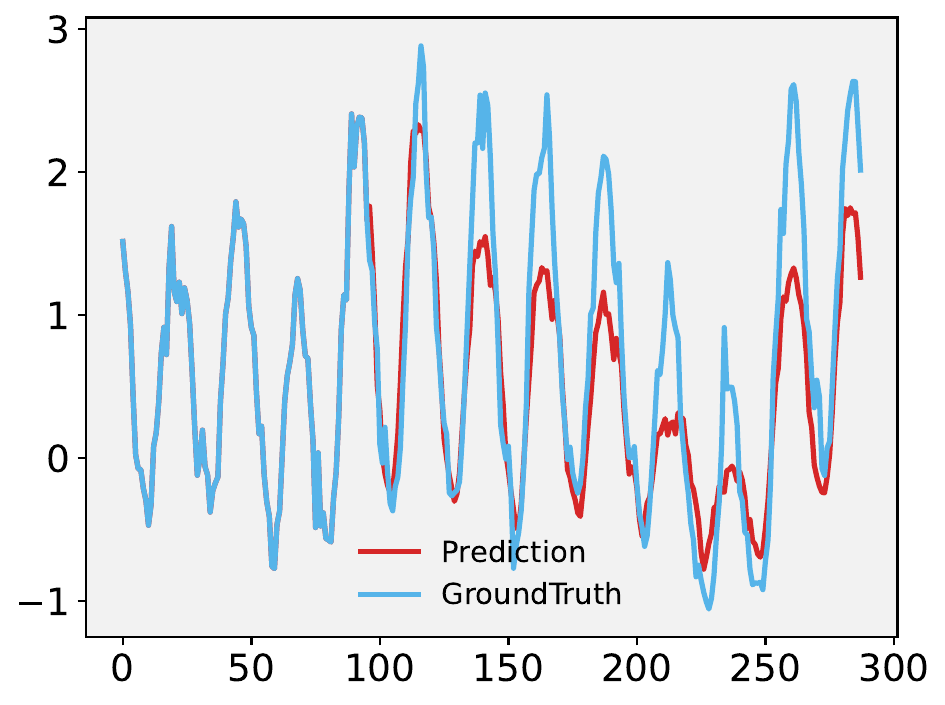}
	}%
 
	\centering
  \vspace{-0.2cm}
    \caption{Visualization of ablation study results on the ECL.}%
	\label{Visualization_abla_photo}
\end{figure}

\subsection{Ablation Studies}
To validate the contribution of each core component in FAiT, we conduct ablation studies on six datasets with the following variants: (i) \textbf{w/o DTFM}, which removes the DTFM module; (ii) \textbf{w/o invAtt}, which removes the Inverted Attention mechanism; (iii) \textbf{Re $\mathbf{Z}_{\text{high}}$}, which substitutes the high-frequency latent $\mathbf{Z}_{\text{high}}$ with the raw basic component $\mathbf{V}$; (iv) \textbf{w/o $\mathcal{W}_\mathcal{M}$}, which disables the parameter sharing strategy between real and imaginary DFT components; and (v) \textbf{w/o ASG}, which employs a single shared fusion weight instead of the adaptive separate gating weights $\mathbf{G}_H$ and $\mathbf{G}_L$.

\begin{figure*}[t]
	\centering
 	\subfloat[FAiT]{
		\includegraphics[scale=0.21]{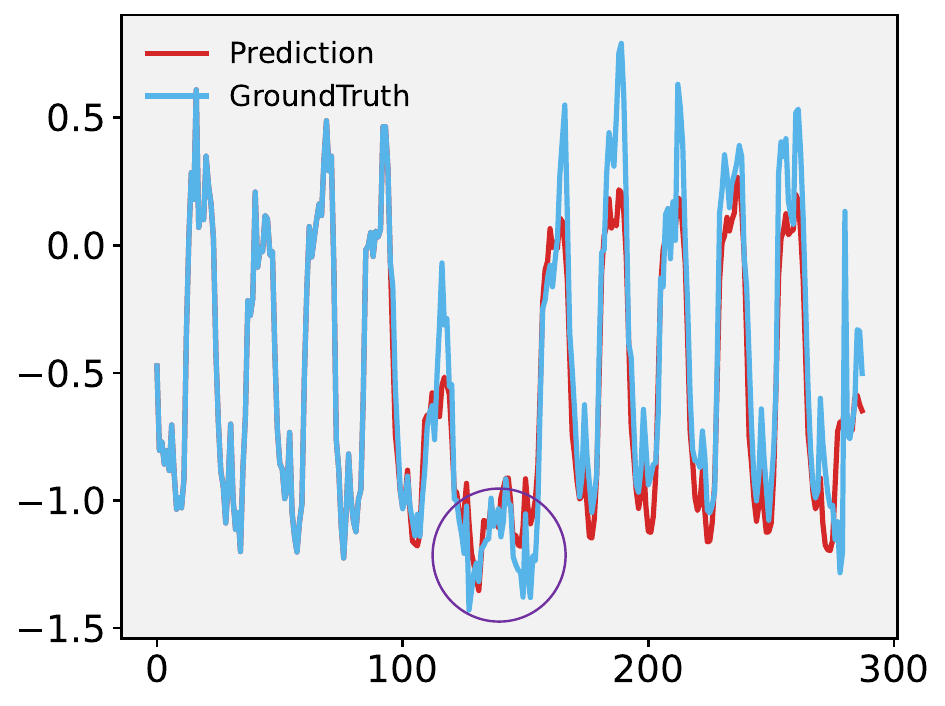}
	}%
 \hspace{-0.1cm}
  	\subfloat[CARD]{
		\includegraphics[scale=0.21]{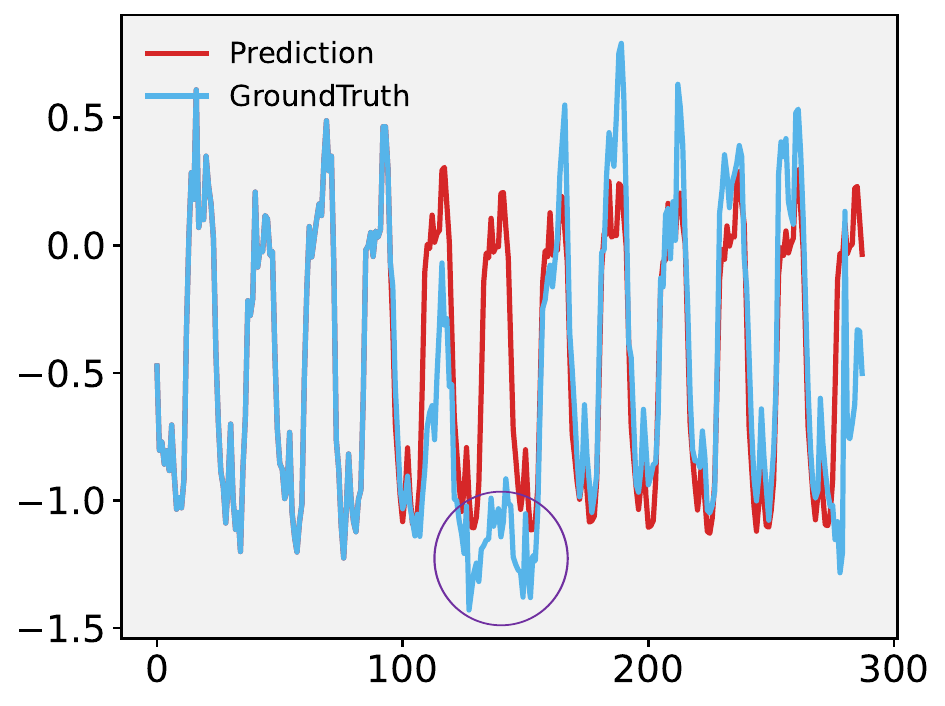}
	}%
 \hspace{-0.1cm}
 	\subfloat[PatchTST]{
		\includegraphics[scale=0.21]{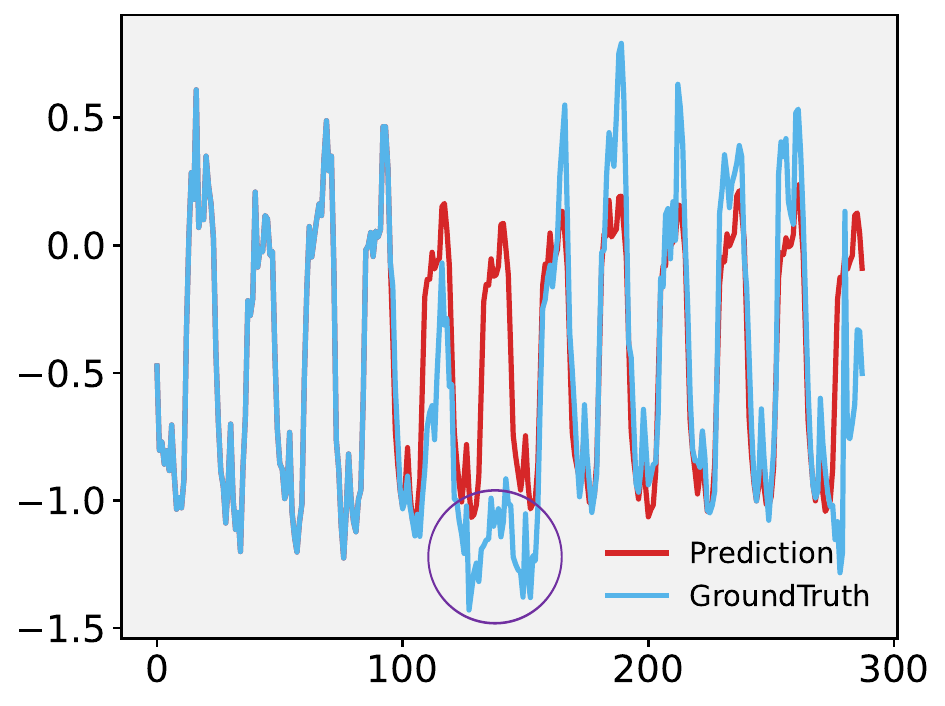}
	}%
 \hspace{-0.1cm}
    \subfloat[FreDF]{
		\includegraphics[scale=0.21]{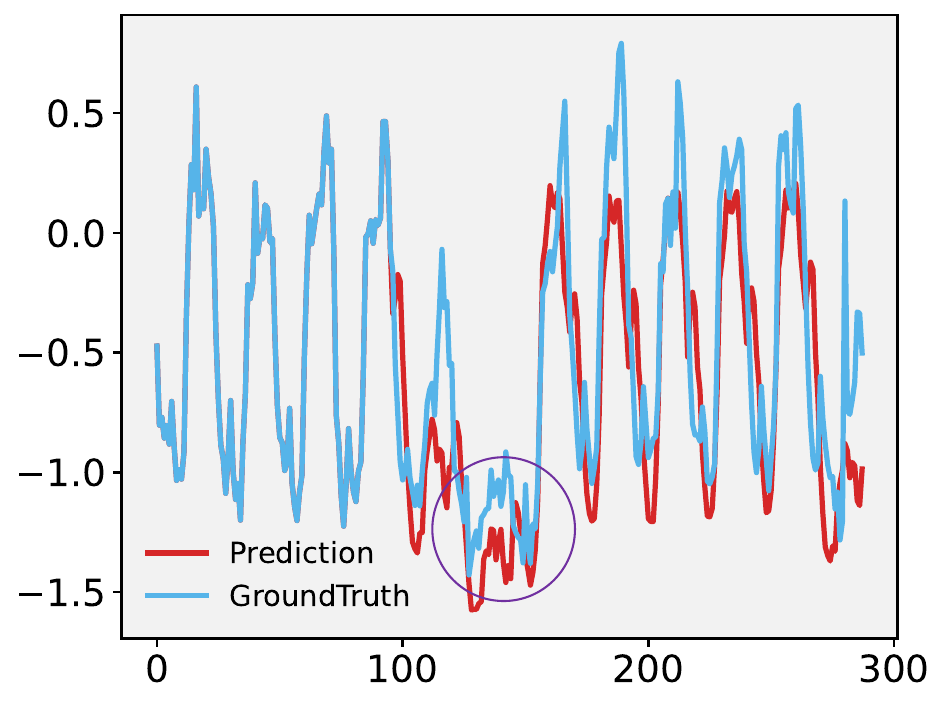}
	}%
 \hspace{-0.1cm}
	   \subfloat[FilterNet]{
		\includegraphics[scale=0.21]{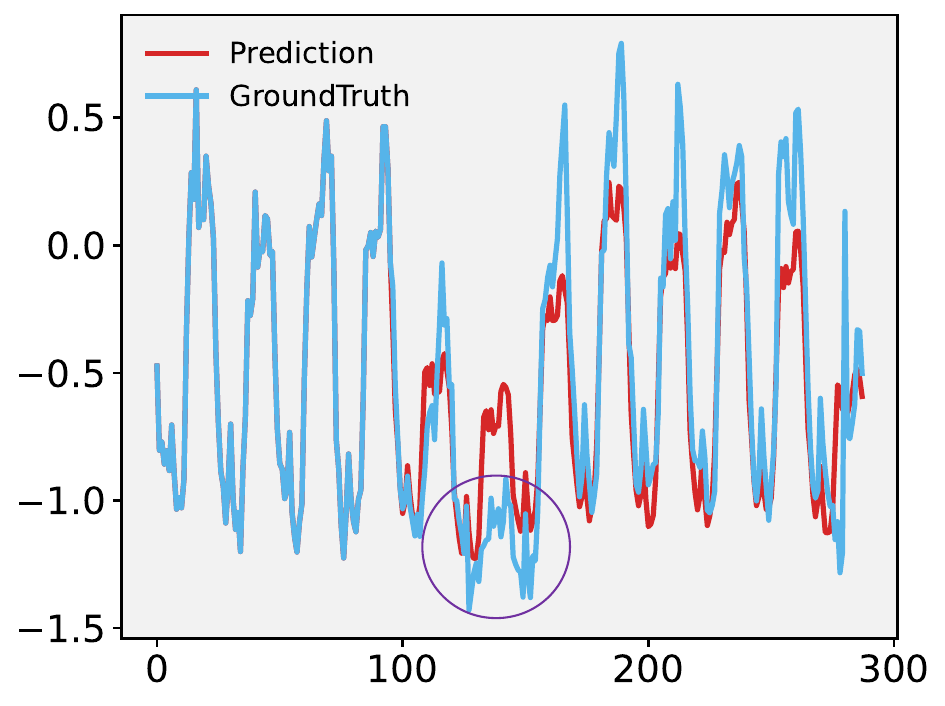}
	}%
	\centering
  \vspace{-0.2cm}
    \caption{Visualization of ECL dataset predictions given by different baselines.}
	\label{Visualization_4}
\end{figure*}

Table \ref{abla} reports the average forecasting performance with a fixed look-back window $L = 96$, evaluated over $H \in \{96, 192, 336, 720\}$. Removing DTFM (\textbf{w/o DTFM}) increases the average MSE/MAE from 0.325/0.359 to 0.334/0.365, confirming the importance of fine-grained, context-dependent reweighting of temporal frequency sub-bands. Similarly, removing Inverted Attention (\textbf{w/o invAtt}) raises the average MSE/MAE to 0.333/0.364, highlighting the benefit of explicitly complementing the low-pass tendency of vanilla attention with high-frequency features. Figure \ref{Visualization_abla_photo} provides a qualitative comparison on ECL (input 96, prediction 192). The \textbf{w/o DTFM} variant exhibits noticeable deviations in fine-grained intervals, while \textbf{w/o invAtt} further amplifies these errors, suggesting that the two modules play complementary roles in stabilizing both local fluctuations and overall trend tracking. We also observe that the non-complementary supplementation strategy (\textbf{Re $\mathbf{Z}_{\text{high}}$}) leads to a clear drop in performance, indicating that aligning high-frequency enhancement with the attention structure is more effective than directly injecting generic high-frequency signals. Furthermore, disabling real-imaginary sharing (\textbf{w/o $\mathcal{W}_\mathcal{M}$}) and using a shared high/low fusion weight (\textbf{w/o ASG}) both degrade performance, validating the design choices for efficient and differentiated fusion of low- and high-frequency streams.

\begin{figure}[t]
	\centering
 	\subfloat[MAE on ETTh1]{
		\includegraphics[scale=0.32]{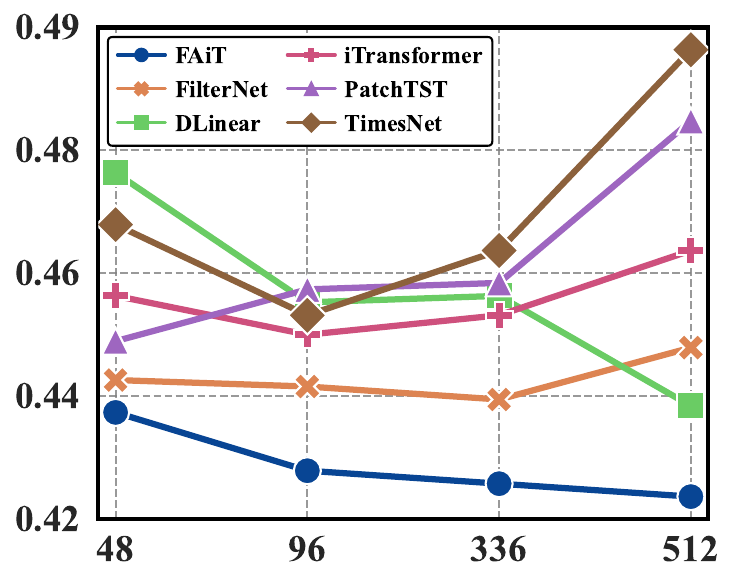}
	}%
 	\subfloat[MAE on ETTm1]{
		\includegraphics[scale=0.32]{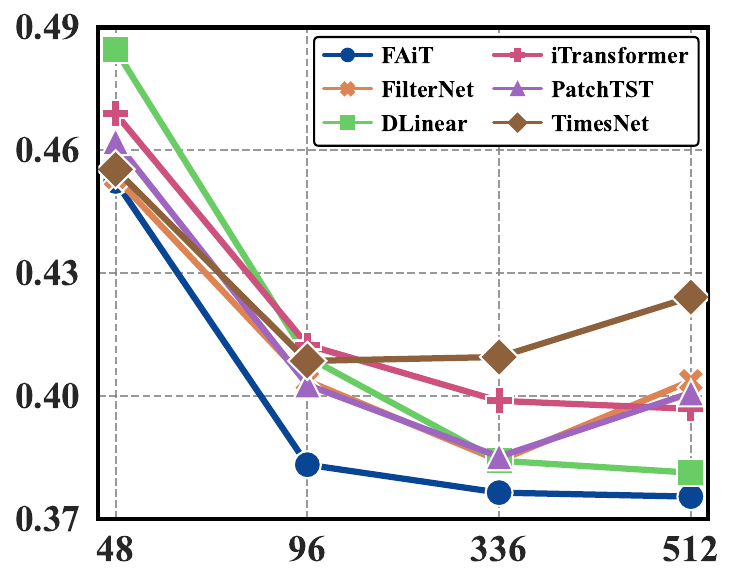}
	}%
	\centering
    \caption{Average performance across look-back lengths $L \in \{48, 96, 336, 512\}$ and horizons $H \in \{96, 192, 336, 720\}$.}
	\label{Visualization_look-back}
\end{figure}

\subsection{Visualization Analysis}
\paragraph{\textbf{Impact of Various Input Length.}}The look-back window size directly affects forecasting accuracy by determining how much historical context a model can exploit. To examine effectiveness under different input lengths, we evaluate model performance under different look-back windows $L \in \{48, 96, 336, 512\}$ on the ETTh1 and ETTm1. Figure \ref{Visualization_look-back} shows that FAiT consistently achieves the lowest MAE across all settings, with errors decreasing steadily as $L$ grows. In contrast, several strong Transformer-based baselines exhibit non-monotonic trends. On ETTh1, iTransformer, PatchTST improve only up to moderate context sizes and then deteriorate noticeably at $L= 512$. A similar trend is also observed on ETTm1. This widening gap under longer inputs aligns with our main claim that vanilla attention acts as a learnable \textbf{low-pass filter} and becomes more prone to \textbf{over-smoothing} as more historical tokens are introduced, reducing sensitivity to fine-grained fluctuations. FAiT circumvents this saturation by explicitly preserving fine-grained dynamics via the Inverted Attention.


 

\paragraph{\textbf{Self-attention Analysis.}}To examine layer-wise feature diversity, we report the average cosine similarity \cite{nguyen2023mitigating} of learned representations for PatchTST, iTransformer, and Transformer in Figure \ref{Visualization_collapse} (more results in Appendix). Higher cosine similarity indicates a stronger tendency toward representation collapse. Across all three backbones, the original models with standard self-attention show a clear monotonic increase in similarity as depth grows, suggesting that stacked layers progressively produce more \textbf{homogeneous features}. After integrating our Inverted Attention into these baselines, the layer-wise similarities are consistently lower and increase more slowly. The difference is most evident in deeper layers, where the self-attention curves continue rising while the corresponding baselines with Inverted Attention remain substantially restrained. This demonstrates that Inverted Attention modules effectively stabilizes the feature manifold and mitigates the severe representation collapse that typically plagues deep Transformer networks.

\begin{figure}[t]
	\centering
     {
		\includegraphics[scale=0.12]{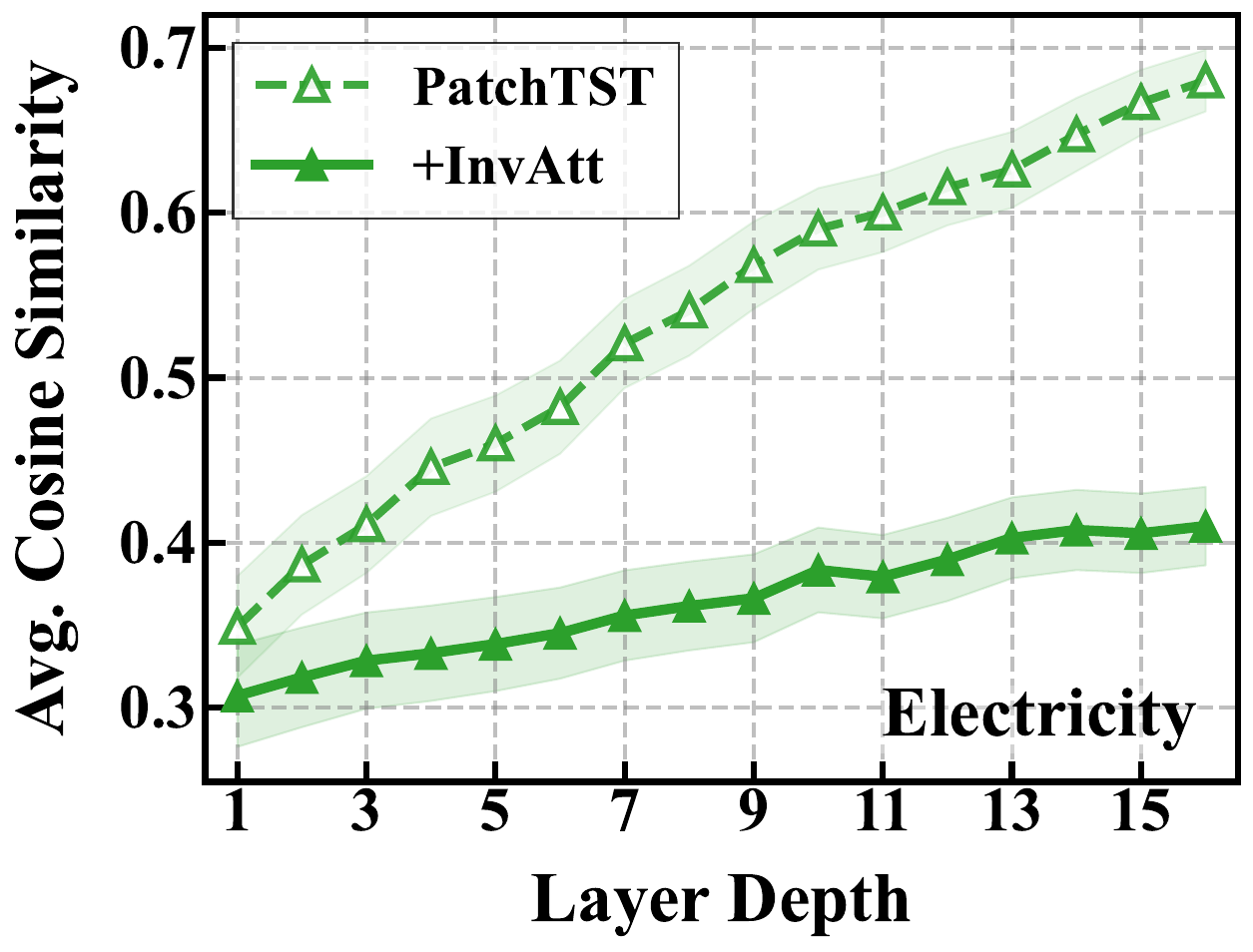}
	}%
	{
		\includegraphics[scale=0.12]{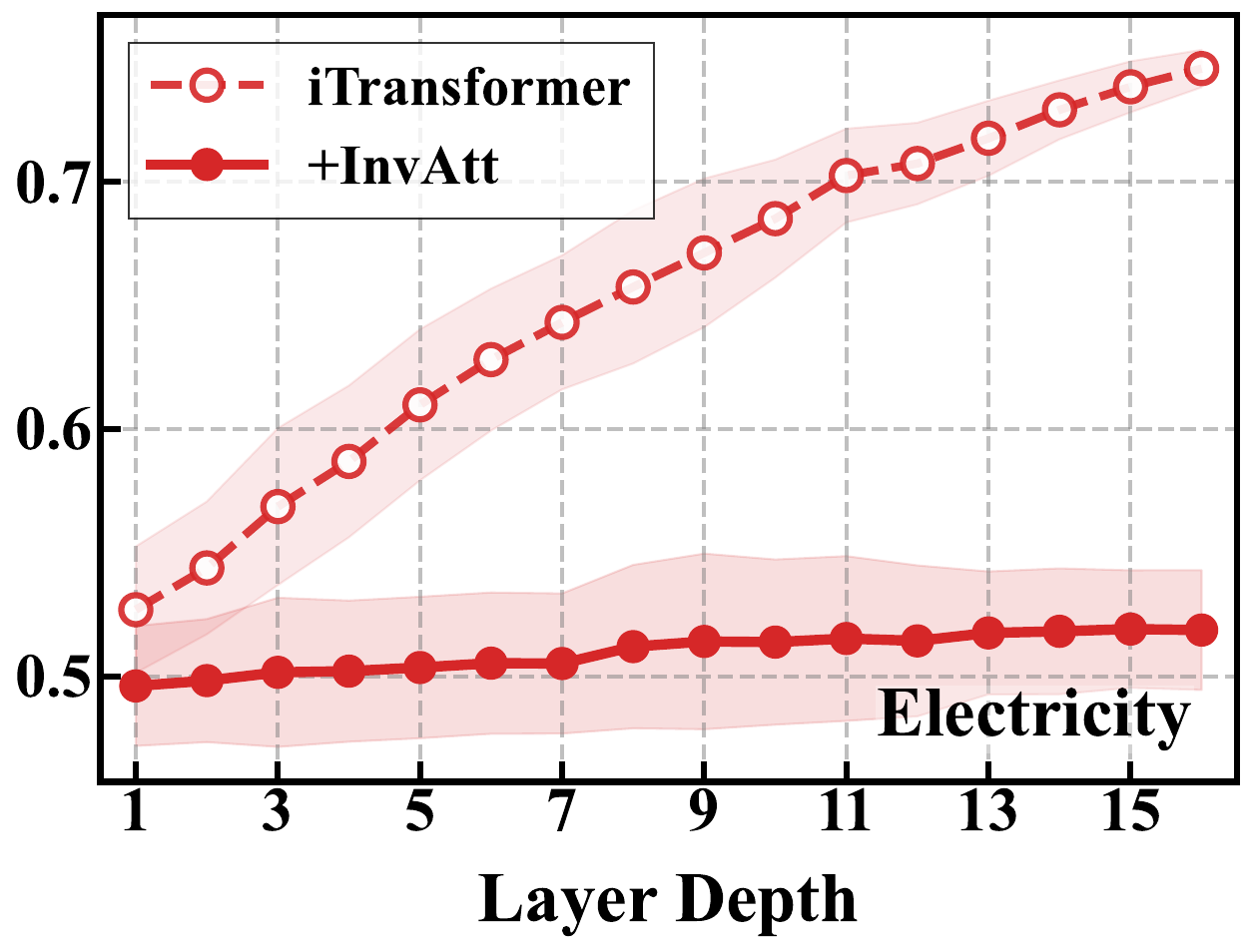}
	}%
    {
		\includegraphics[scale=0.12]{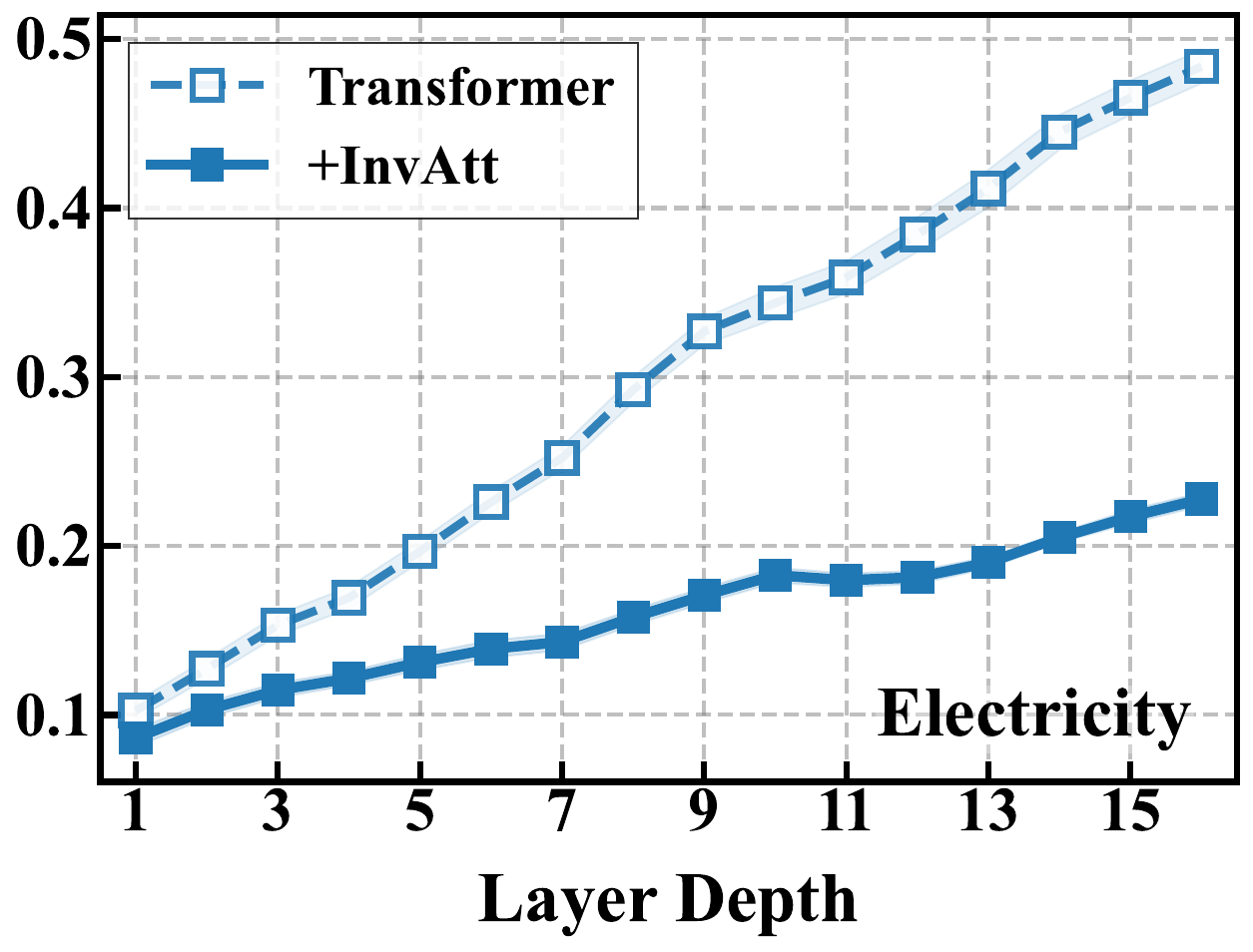}
	}%
     \\[-0.4cm]
     
     \subfloat[PatchTST]{
		\includegraphics[scale=0.12]{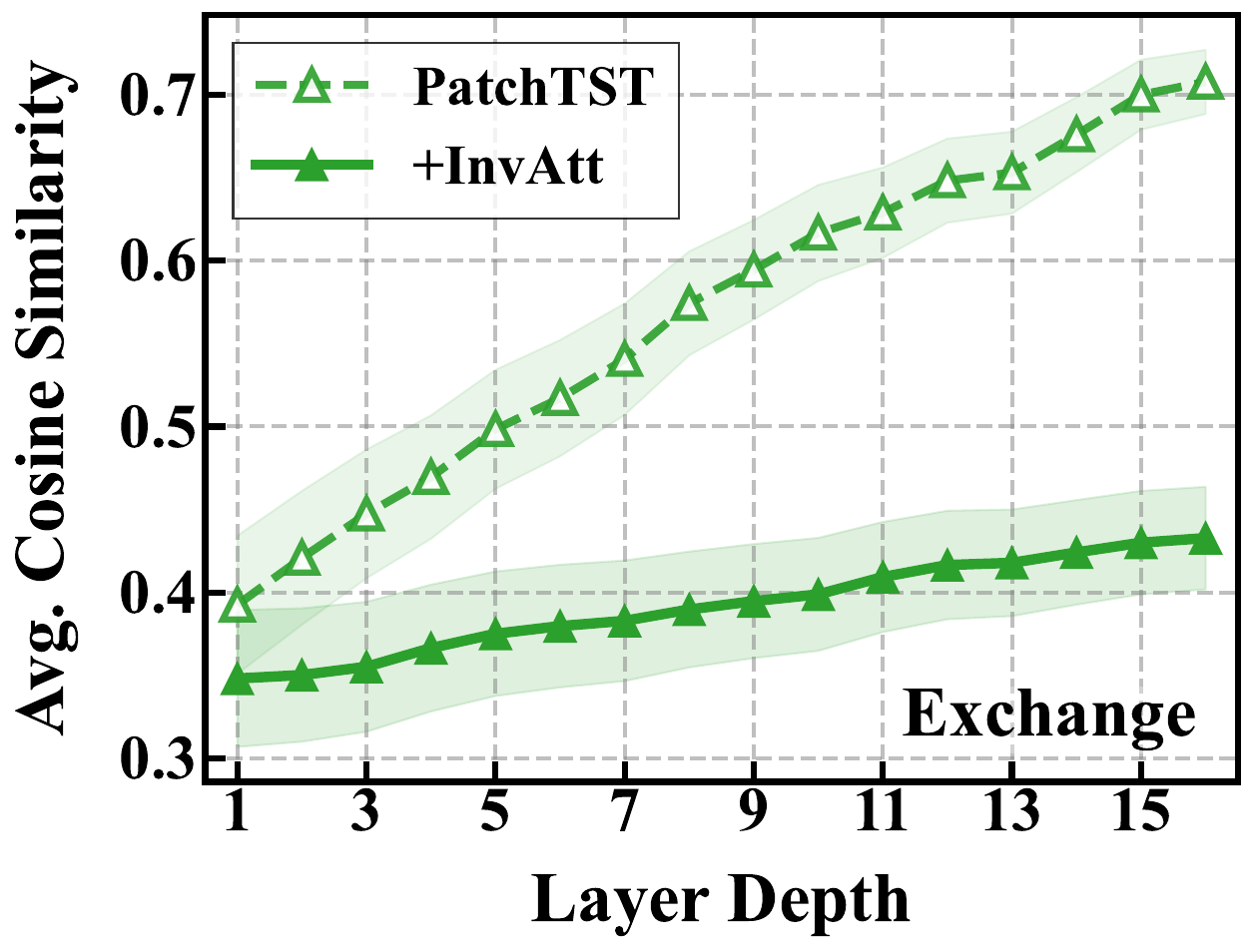}
	}%
	   \subfloat[iTransformer]{
		\includegraphics[scale=0.12]{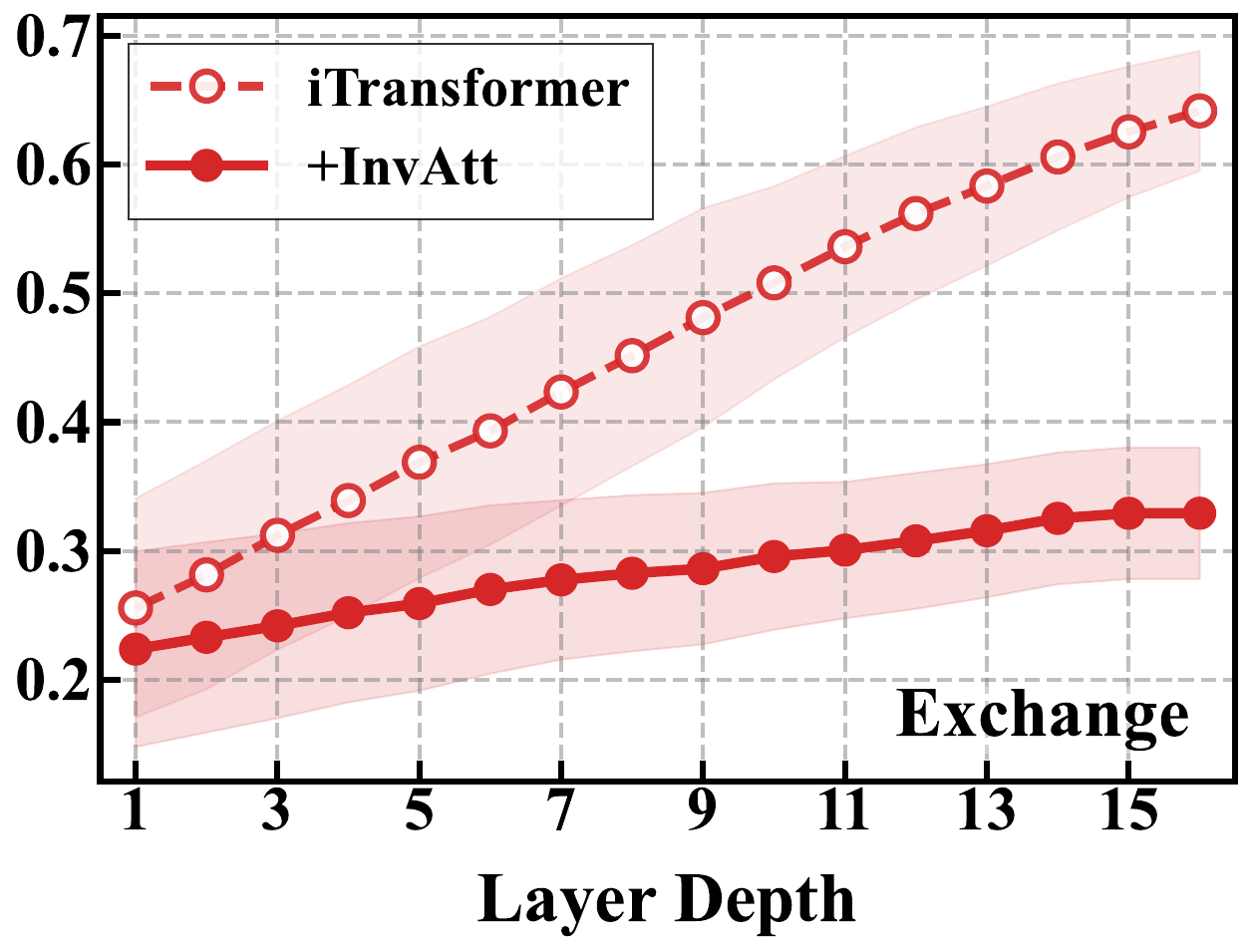}
	}%
    \subfloat[Transformer]{
		\includegraphics[scale=0.12]{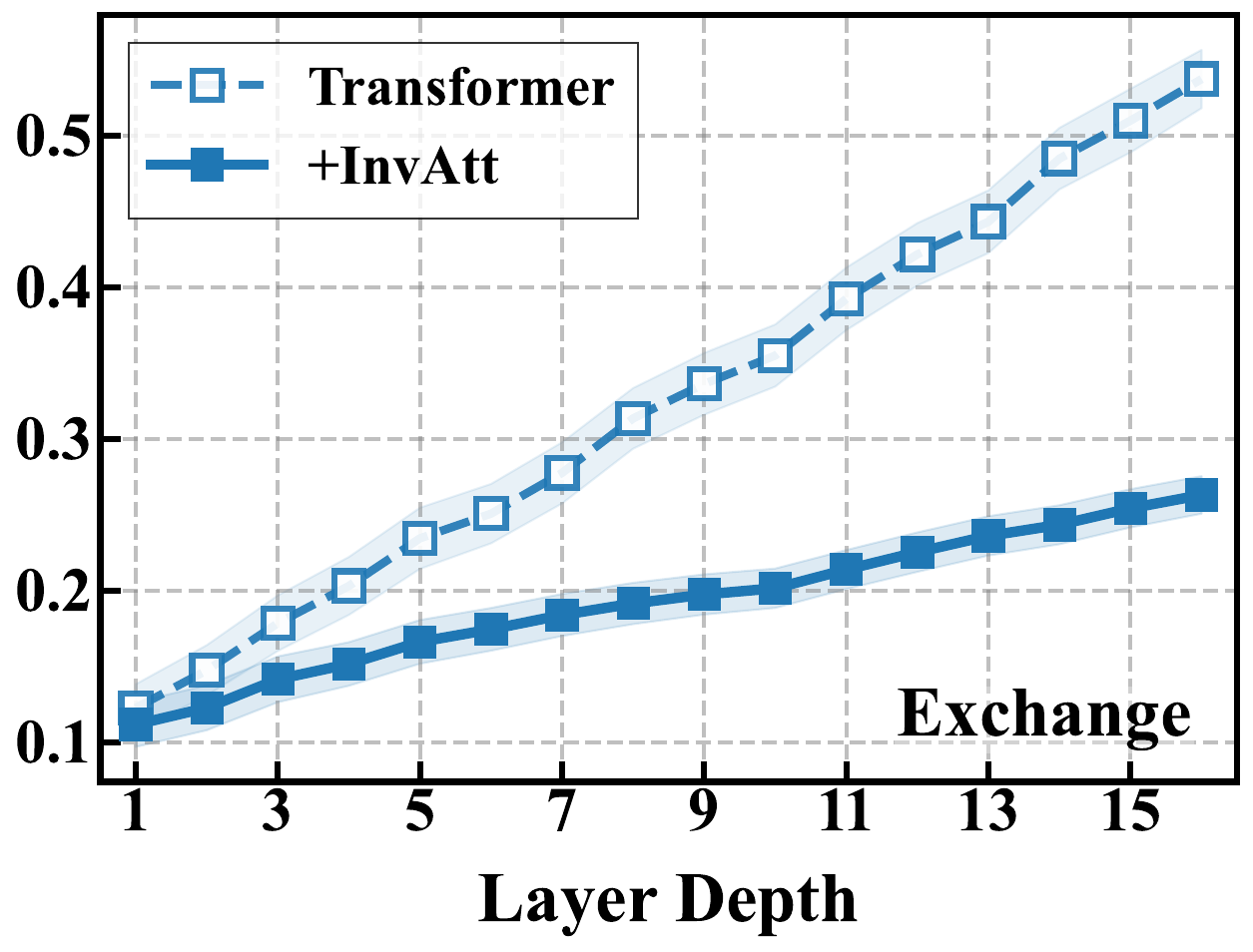}
	}%
	\centering
 \vspace{-0.2cm}
    \caption{Comparison of layer-wise representation collapse. 
    }
	\label{Visualization_collapse}
\end{figure}



\paragraph{\textbf{Qualitative Analysis of Non-Stationarity.}}

To demonstrate FAiT's capability in modeling complex, non-stationary dynamics, we conduct a qualitative analysis on the ECL dataset (Figure \ref{Visualization_4}). The circled regions highlight a brief disruption of an otherwise regular oscillatory pattern. In these segments, standard attention-based baselines (CARD and PatchTST) produce visibly smoother responses and underestimate the abrupt drop, suggesting that their intrinsic low-pass tendency weakens sensitivity to rapid, transient fluctuations. Frequency-enhanced baselines (FreDF and FilterNet) appear more responsive to oscillatory structure, yet still show noticeable deviations around the turning point in the highlighted region, with less accurate matching of the local drop-and-recovery behavior. In contrast, FAiT follows the disrupted segment more closely and aligns better with the subsequent oscillations. This visual evidence verifies that explicitly introducing a complementary high-pass branch via Inverted Attention, together with band-wise refinement through DTFM, helps preserve fine-grained temporal variations without sacrificing overall trend consistency.

\begin{figure}[t]
	\centering
 	\subfloat[Spectral Responses on ETTh2]{
		\includegraphics[scale=0.21]{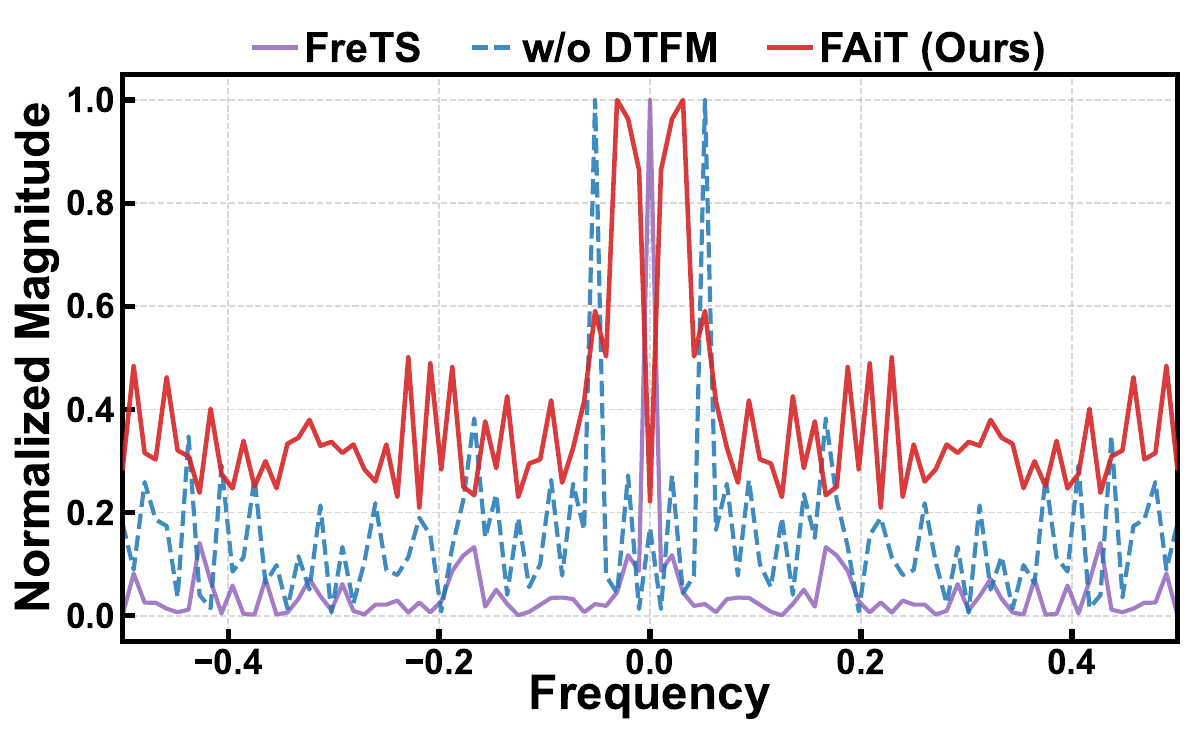}
	}
 	\subfloat[Spectral Responses on ECL]{
		\includegraphics[scale=0.21]{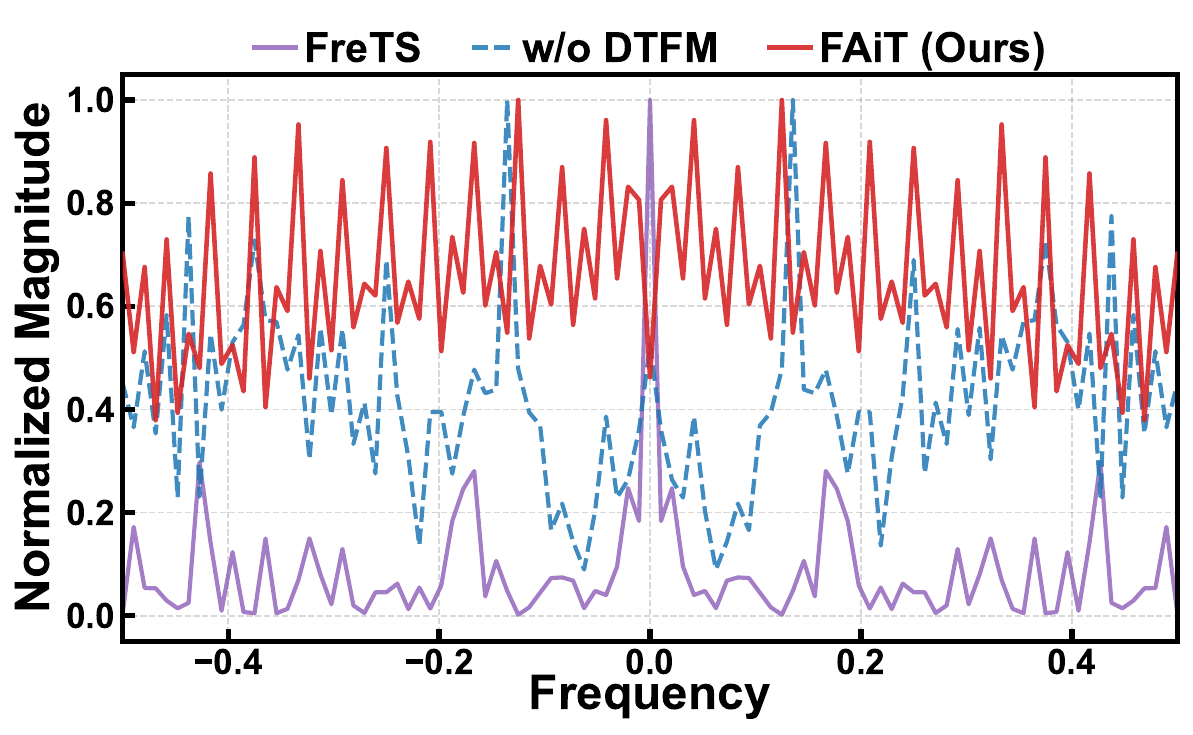}
	}%
	\centering
    \caption{Visualization of spectral responses for different filters on ETTh2 and ECL datasets.}
	\label{Visualization_Spectral}
\end{figure}

\paragraph{\textbf{Spectral Responses of Filters.}} 
To validate the effectiveness of DTFM, we visualize the learned spectral responses\footnote{Visualization implementation adapted from BSARec \cite{shin2024attentive}. Source code: \url{https://github.com/yehjin-shin/BSARec/blob/main/src/visualize/figure2.ipynb}.} on the ETTh2 and ECL datasets in Figure \ref{Visualization_Spectral}. As illustrated, frequency-domain baselines (e.g., FreTS, \textcolor{violet}{purple} line) still exhibit noticeable spectral decay, predominantly favoring smooth low-frequency components. In contrast, the ablation variant w/o DTFM (\textcolor{blue}{dashed} line) successfully mitigates this bias, restoring a substantial portion of critical high-frequency energy via the Inverted Attention mechanism. Building on this foundation, FAiT (with DTFM, \textcolor{red}{solid} line) further elevates performance by exhibiting significantly sharper and more distinct spectral peaks. These active responses demonstrate that DTFM goes beyond merely passing high-frequency signals; it leverages the restored energy for adaptive, fine-grained spectral recalibration, thereby capturing evolving complex, instance-specific variations that static coarse inversion alone cannot fully resolve.

\subsection{Sensitivity Analysis}

\begin{figure}[t]
	\centering
   \subfloat[$D$ on ETTm1]{
		\includegraphics[scale=0.14]{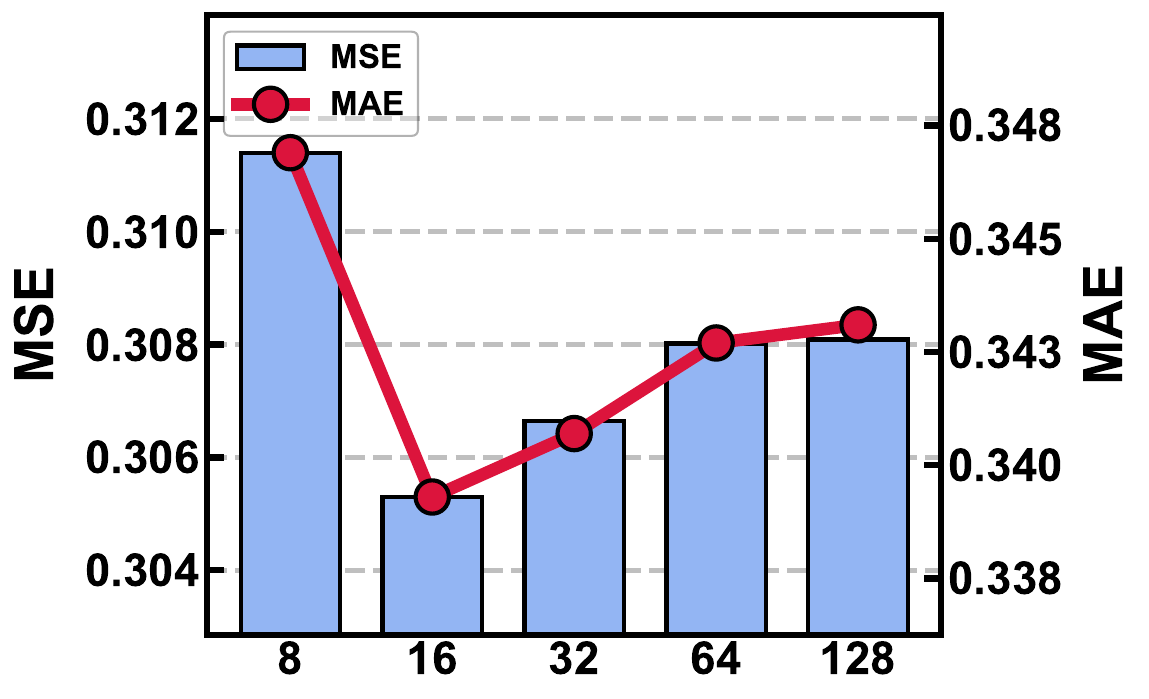}
	}%
 \hspace{-0.2cm}
 	\subfloat[$G$ on ETTm1]{
		\includegraphics[scale=0.14]{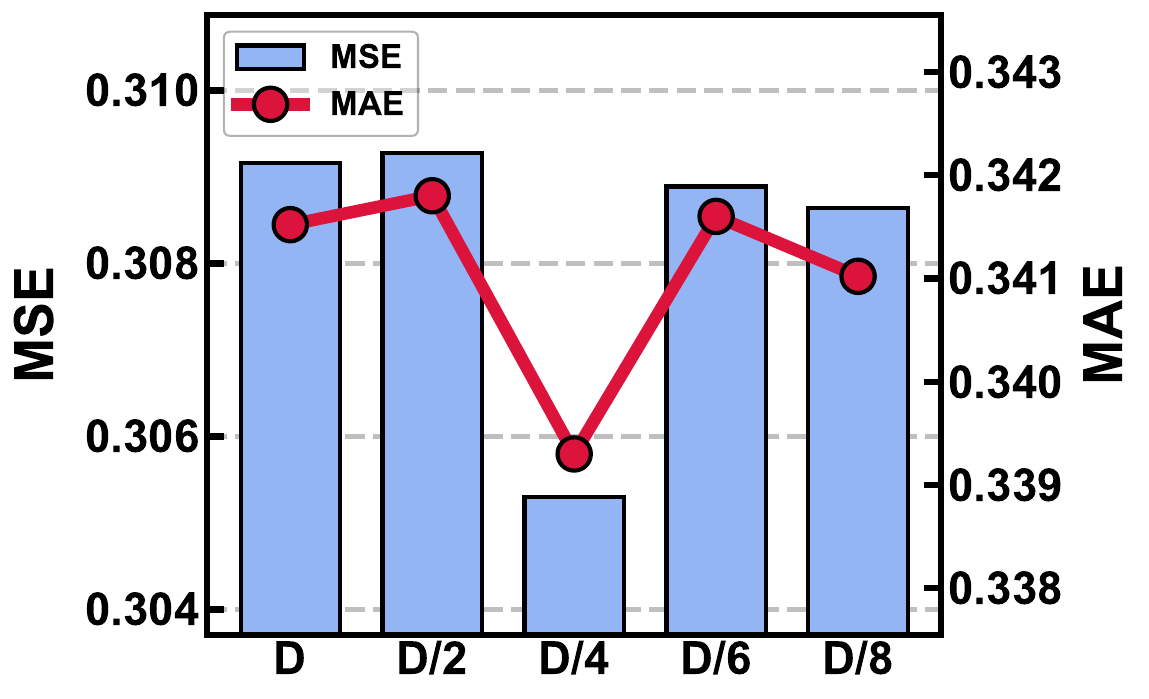}
	}%
 \hspace{-0.2cm}
  	\subfloat[$F$ on ETTm1]{
		\includegraphics[scale=0.14]{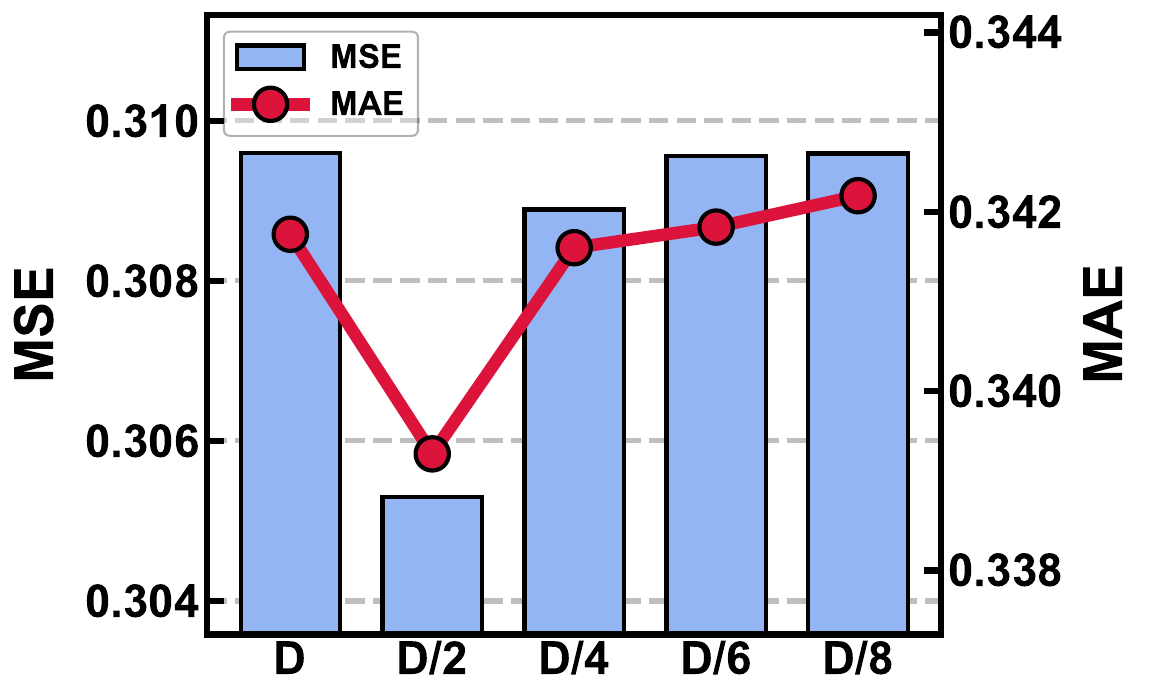}
	}%
 \vspace{-1em}
 	\subfloat[$D$ on Weather]{
		\includegraphics[scale=0.14]{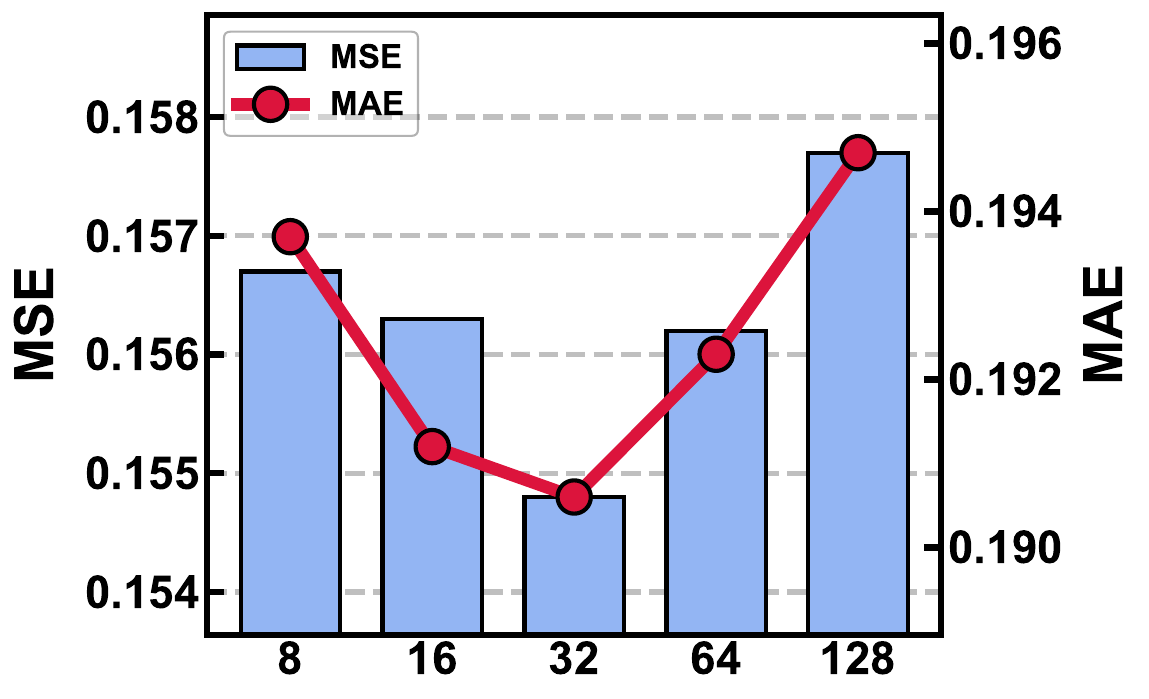}
	}%
 \hspace{-0.2cm}
 	\subfloat[$G$ on Weather]{
		\includegraphics[scale=0.14]{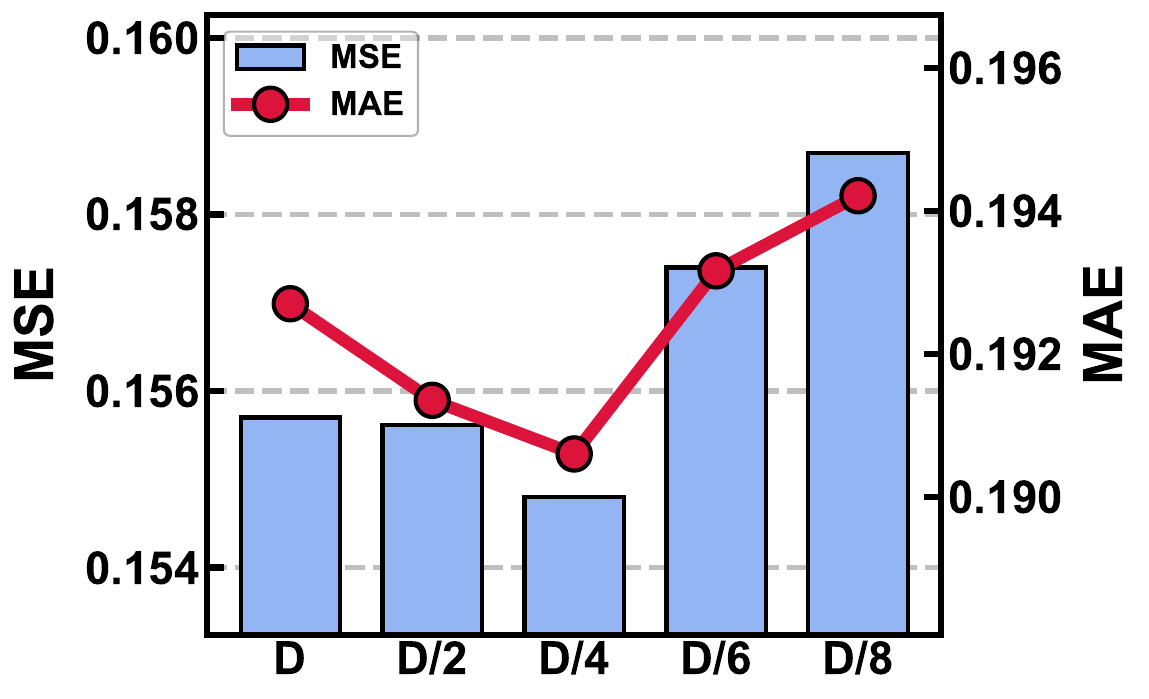}
	}%
 \hspace{-0.2cm}
 	\subfloat[$F$ on Weather]{
		\includegraphics[scale=0.14]{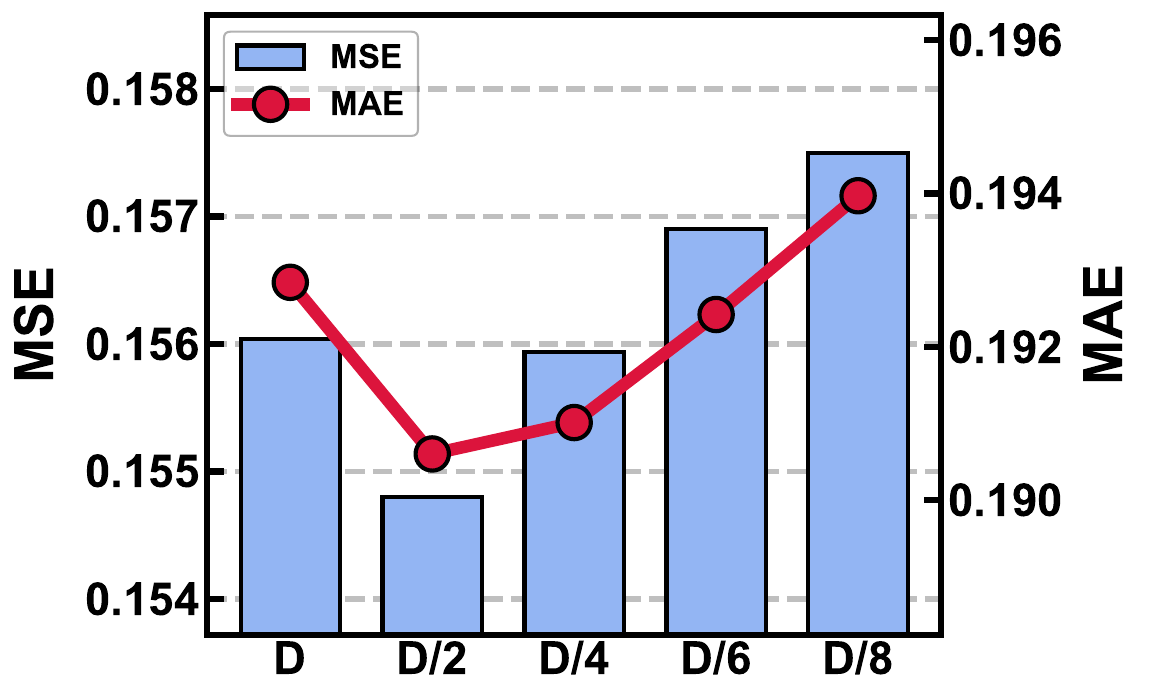}
	}
	\centering
    \caption{Sensitivity analysis of Embedding size $D$, group $G$ and filter $F$ on ETTm1 and Weather datasets ($L=H=96$).}
	\label{Visualization_3}
\end{figure}

We analyze the sensitivity of FAiT to embedding dimension $D$, group size $G$, and filter count $F$ on ETTm1 and Weather (Figure \ref{Visualization_3}).
\begin{itemize}[leftmargin=*]
    \item \noindent\textbf{Impact of Dimension $D$.} As shown in Figures \ref{Visualization_3} (a, d), the MSE/MAE metrics exhibit a ‘U’-shaped trend, reaching optimal performance at $D=16$ (for ETTm1) and $D=32$ (for Weather). Extreme values degrade results due to either insufficient model capacity or increased optimization difficulty.
    \item \noindent\textbf{Impact of Structure ($G$ and $F$).} As shown in Figures \ref{Visualization_3} (b, e) and (c, f), both parameters show similar ‘U’-shaped patterns, consistently achieving optima at $G = D/4$ and $F = D/2$ on both datasets. This indicates that a balanced structural allocation is crucial; excessive groups or filters unnecessarily increase learning redundancy, while insufficient numbers limit feature extraction, both of which impair final prediction accuracy.
\end{itemize}

\subsection{Efficiency Analysis}

\begin{figure}[t]
  \centering
  \includegraphics[width=0.7\linewidth]{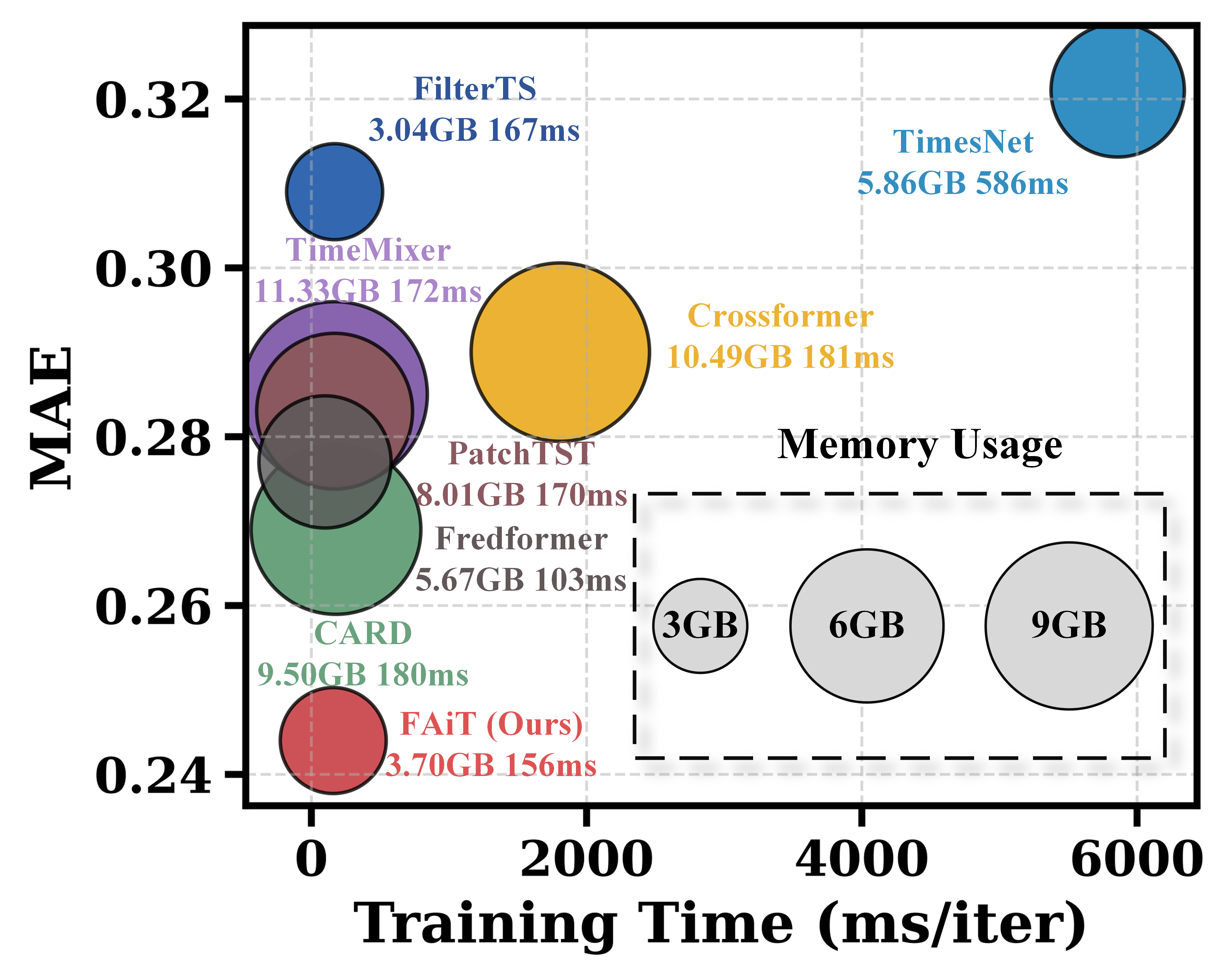}
   \vspace{-0.2cm}
  \caption{Training memory usage (GB), training time (ms/iter), and MAE comparisons on the Traffic dataset. The input and predicted lengths are set to 96 and 96, respectively.}
  \label{Efficiency}
\end{figure}

Figure \ref{Efficiency} illustrates the efficiency-error trade-off on Traffic dataset, where FAiT occupies the optimal bottom-left region, achieving superior forecasting performance with the lowest MAE and a relatively memory footprint of \textbf{3.70 GB}. Specifically, our model outperforms the leading Transformer-based baseline CARD (9.50 GB) and the MLP-based baseline TimeMixer (11.33 GB) in memory efficiency by approximately $\mathbf{2.6\times}$ and $\mathbf{3\times}$, respectively, while maintaining a highly competitive training speed of \textbf{156 ms/iter}. Furthermore, compared to the frequency-domain baseline Fredformer (5.67 GB), FAiT leverages a streamlined Inverted Attention mechanism to maintain high fidelity without additional parameters, yielding a lower MAE. 

\section{Conclusion}
In this paper, we present FAiT, a Frequency-Aware Inverted Transformer designed to rectify the inherent low-pass filtering bias of self-attention in MTSF. Unlike existing frequency-enhanced approaches that often rely on static bases or global projections, FAiT introduces a novel dual-path Inverted Attention mechanism to explicitly decouple and preserve vital high-frequency information, thereby recovering rapid fluctuations typically suppressed by vanilla attention. Furthermore, by incorporating Dynamic Temporal-Frequency Modulation, our model overcomes the limitations of coarse-grained, sequence-wise spectral operations. This enables instance-adaptive spectral calibration that dynamically reweights frequency bands to match complex evolving non-stationary patterns. Extensive experiments on widely-used benchmarks validate that FAiT effectively captures both global trends and fine-grained local transients, consistently outperforming competitive state-of-the-art methods.

\newpage
\bibliographystyle{ACM-Reference-Format}
\bibliography{Reference}

\appendix
\newpage

\section{Calculation of Spectral Energy Preservation Ratio}
\label{appendix:spectral_ratio}

To quantitatively evaluate the extent to which different models preserve the intrinsic spectral properties of time series, particularly to measure the mitigation of spectral attenuation caused by low-pass filtering bias in self-attention mechanisms, we introduce the \textit{Spectral Energy Preservation Ratio} \cite{oppenheim1999discrete}.

Let $\mathbf{x} \in \mathbb{R}^{L}$ denote a segment of the ground-truth time series, and $\hat{\mathbf{x}} \in \mathbb{R}^{L}$ denote the corresponding output generated by the model. We first transform both signals from the time domain to the frequency domain using the Discrete Fourier Transform:
\begin{equation}
    \mathcal{X}[k] = \sum_{n=0}^{L-1} \mathbf{x}[n] e^{-j \frac{2\pi}{L} k n}, \quad \hat{\mathcal{X}}[k] = \sum_{n=0}^{L-1} \hat{\mathbf{x}}[n] e^{-j \frac{2\pi}{L} k n},
\end{equation}
where $k = 0, 1, \dots, L-1$ represents the frequency component index, and $L$ is the sequence length.

The amplitude spectra of the ground truth and the model output are derived as the magnitudes of their respective complex spectral coefficients:
\begin{equation}
    S_{target}[k] = \left| \mathcal{X}[k] \right|, \quad S_{model}[k] = \left| \hat{\mathcal{X}}[k] \right|.
\end{equation}

Consistent with the principle of Parseval's theorem \cite{parseval1806memoire}, the spectral energy is proportional to the sum of the squared amplitudes. We define the total spectral energy for the target and the model within the analyzed frequency range $\mathcal{K}$ as:
\begin{equation}
    E_{target} = \sum_{k \in \mathcal{K}} \left( S_{target}[k] \right)^2, \quad E_{model} = \sum_{k \in \mathcal{K}} \left( S_{model}[k] \right)^2.
\end{equation}

Finally, the Spectral Energy Preservation Ratio ($\eta$) is defined as the percentage of the ground-truth spectral energy retained by the model's output:
\begin{equation}
    \eta = \frac{E_{model}}{E_{target}} \times 100\% = \frac{\sum_{k \in \mathcal{K}} \left| \hat{\mathcal{X}}[k] \right|^2}{\sum_{k \in \mathcal{K}} \left| \mathcal{X}[k] \right|^2} \times 100\%.
\end{equation}

A ratio $\eta \approx 100\%$ indicates that the model effectively preserves the energy distribution of the original signal. A ratio $\eta < 100\%$ (or even $< 50\%)$ signifies spectral loss, which is characteristic of the low-pass filtering effect observed in vanilla Transformers. 

\section{Theoretical Analysis of FAiT}
\label{appendix:theoretical_analysis}

We provide a Graph Signal Processing (GSP)-based justification for \textbf{Inverted Attention}. We demonstrate that while self-attention acts as a low-pass filter, our inverted branch $(\mathbf{I} - \mathcal{A})$ functions as a high-pass filter, effectively recovering local irregularities. 

\subsection{Self-Attention as Graph Signal Processing}
We view the input $V \in \mathbb{R}^{L \times D}$ as a multivariate graph signal defined on $\mathcal{G}$, where the node set $\mathcal{V}$ corresponds to the $L$ time steps of the input. The normalized attention map $\mathcal{A}$ is:
\begin{equation}
    \mathcal{A} = \text{Softmax}\left(\frac{Q K^\top}{\sqrt{D}}\right).
\end{equation}
In the context of GSP, $\mathcal{A}$ serves as the \textit{row-stochastic adjacency matrix} \cite{ortega2018graph,wang22anti,choi2024graph}. Consequently, $\mathbf{Z}_{\text{low}} = \mathcal{A}V$ represents a single-step diffusion or localized averaging operation.


\begin{algorithm}[t]
\caption{PyTorch-style pseudocode for Inverted Transformer}
\label{alg:inverted_attention}
\begin{lstlisting}[language=Python]
# d: embedding dimension
gate_L  = nn.Linear(d, d) 
gate_H  = nn.Linear(d, d)
proj    = nn.Linear(d, d)

def InvertedTransformer(x, attn_layer):
    # x: Input tensor [B, C, T, D]
    
    # 1. Project to Q, K, V
    q, k, v = split(qkv_proj(x)) 
    
    # 2. Low-Frequency Stream (Standard Attention)
    # The standard attention acts as a low-pass filter
    z_low = attn_layer(q, k, v)
    
    # 3. High-Frequency Stream (Inverted Branch)
    # Recover high-freq by Inverted Attention
    z_high = v - z_low
    
    # 4. Adaptive Spectral Gating (ASG)
    # Compute dynamic gates for frequency bands
    g_low  = gate_L(x).tanh()
    
    # Softplus-based gating for high-freq stability
    g_high_raw = F.softplus(gate_H(x))
    g_high = 2 * (g_high_raw**2) / (g_high_raw**2 + 0.3678)
    
    # 5. Adaptive Fusion
    out =g_low * z_low + g_high * z_high
                
    return proj(out)
\end{lstlisting}
\end{algorithm}


    
    
    
    
    
    
    

\subsection{Spectral Analysis of Attention Filters}
Let $\{\lambda_i\}_{i=1}^L$ be the eigenvalues of $\mathcal{A}$ ($|\lambda_i| \le 1$). Standard attention \cite{vaswani2017attention} aggregates similar neighbors, corresponding to a spectral filter response $h_{low}(\lambda) = \lambda$.
This operation tends to minimize the \textit{Dirichlet Energy}, retaining global trends associated with $\lambda_i \approx 1$ while attenuating high-frequency oscillations where $\lambda_i$ is small. Thus, the attention matrix $\mathcal{A}$ acts as a low-pass filter, causing the over-smoothing problem in deep Transformers.

\subsection{Inverted Attention as High-pass Filtering}
To restore intrinsic high-frequency information that is typically suppressed, we propose an \textbf{Inverse Attention} mechanism:
\begin{equation}
    \mathcal{L}_{rw} = \mathbf{I} - \mathcal{A}.
\end{equation}
Here, $\mathbf{I}$ is the identity matrix, and our inverted branch calculation is further expressed as $\mathbf{Z}_{\text{high}} = (\mathbf{I} - \mathcal{A})\mathbf{V} = \mathcal{L}_{rw} \mathbf{V}$.


\textbf{Theorem 1 (High-Pass Property of Inverted Attention).} \textit{The Inverted Attention operator $\mathcal{L}_{rw} = \mathbf{I} - \mathcal{A}$ functions as a high-pass graph filter that maximizes the variation of the output signal relative to the graph structure defined by $\mathcal{A}$.}

\textit{Proof.}
Consider the spectral decomposition of the operator. Since the eigenvalues of $\mathcal{A}$ are $\lambda_i$, the eigenvalues of $\mathcal{L}_{rw}$ are given by $\mu_i = 1 - \lambda_i$.
The frequency response of the Inverted Attention is thus $h_{high}(\lambda) = 1 - \lambda$.
\begin{itemize}[leftmargin=*]
    \item \textbf{Low Frequencies:} For slowly varying components (global trends), the corresponding eigenvalues of $\mathcal{A}$ are close to 1 ($\lambda \approx 1$). The response of the inverted branch is $\mu \approx 1 - 1 = 0$. Thus, global trends are attenuated.
    \item \textbf{High Frequencies:} For rapidly changing components (local irregularities), the eigenvalues of $\mathcal{A}$ are small ($\lambda \to 0$ or negative). The response of the inverted branch is $\mu \approx 1 - 0 = 1$. Thus, high-frequency details are preserved and highlighted.
\end{itemize}

Specifically, $\mathcal{L}_{rw} \mathbf{V}$ computes the local residual:
\begin{equation}
    (\mathcal{L}_{rw} \mathbf{V})_i = \mathbf{V}_i - \sum_{j} \mathcal{A}_{ij} \mathbf{V}_j.
\end{equation}
This captures local variations orthogonal to the smoothing effect. Therefore, FAiT's dual-branch design ($\mathbf{Z} = \mathbf{G}_L \odot \mathbf{Z}_{\text{low}} + \mathbf{G}_H \odot \mathbf{Z}_{\text{high}}$) constructs an all-pass filter bank: $\mathcal{A}$ extracts trends (minimizing Dirichlet energy), while $(\mathbf{I}-\mathcal{A})$ captures volatility (maximizing it), effectively preventing spectral collapse.

\begin{table*}[t]
\centering
\fontsize{9pt}{9pt} \selectfont 
\renewcommand{\arraystretch}{0.8} 
\setlength{\tabcolsep}{2.9pt} 
\caption{Comparison of FAiT and baseline models on five real-world datasets for an input sequence length of $L = 96$ and a forecast horizon of $H = 96$. The comparison includes model parameters, FLOPs, training/inference time (T.T./I.T.), and training/inference memory (T.M./I.M.). All models were run with a batch size of 16, with remaining parameters set as per their original implementations.}
\begin{tabular}{@{}cc|ccccccccccc@{}}
\toprule
\multicolumn{2}{c|}{Models} & \begin{tabular}[c]{@{}c@{}}FAiT\\ (Ours)\end{tabular} & \begin{tabular}[c]{@{}c@{}}FilterTS\\ \citeyearpar{wang2025filterts}\end{tabular} & \begin{tabular}[c]{@{}c@{}}CARD\\ \citeyearpar{wang2024card}\end{tabular} & \begin{tabular}[c]{@{}c@{}}Crossfm.\\ \citeyearpar{zhang2023crossformer}\end{tabular} & \begin{tabular}[c]{@{}c@{}}iTrans.\\ \citeyearpar{liu2024itransformer}\end{tabular} & \begin{tabular}[c]{@{}c@{}}Fredfm.\\ \citeyearpar{piao2024fredformer}\end{tabular} & \begin{tabular}[c]{@{}c@{}}TimeMixer\\ \citeyearpar{wang2023timemixer}\end{tabular} & \begin{tabular}[c]{@{}c@{}}DLinear\\ \citeyearpar{zeng2023transformers}\end{tabular} & \begin{tabular}[c]{@{}c@{}}TimesNet\\ \citeyearpar{wu2023timesnet}\end{tabular} & \begin{tabular}[c]{@{}c@{}}PatchTST\\ \citeyearpar{nie2023a}\end{tabular} & \begin{tabular}[c]{@{}c@{}}FreTS\\ \citeyearpar{yi2023frequency}\end{tabular} \\ \midrule
\multicolumn{1}{c|}{\multirow{6}{*}{ETTh1}} & Params(M) & 5.57 & 0.68 & 0.03 & 36.97 & 4.83 & 8.59 & 0.08 & 0.02 & 299.94 & 3.76 & 0.42 \\
\multicolumn{1}{c|}{} & FLOPs(M) & 51.16 & 8.13 & 1.41 & 1,507.47 & 33.99 & 135.01 & 10.72 & 0.13 & 289,708.00 & 272.20 & 2.94 \\
\multicolumn{1}{c|}{} & T.T.(ms/iter) & 23.53 & 26.60 & 29.10 & 90.05 & 10.11 & 26.48 & 27.99 & 1.42 & 501.98 & 8.37 & 3.63 \\
\multicolumn{1}{c|}{} & T.M.(GB) & 0.12 & 0.05 & 0.03 & 1.11 & 0.21 & 0.24 & 0.05 & 0.02 & 5.80 & 0.14 & 0.03 \\
\multicolumn{1}{c|}{} & I.T.(ms/iter) & 4.34 & 9.74 & 4.05 & 24.90 & 1.72 & 3.73 & 3.86 & 0.21 & 155.97 & 1.38 & 0.49 \\
\multicolumn{1}{c|}{} & I.M.(MB) & 34.69 & 20.38 & 11.02 & 201.63 & 156.00 & 78.04 & 18.06 & 8.48 & 1,396.24 & 51.83 & 14.47 \\ \midrule
\multicolumn{1}{c|}{\multirow{6}{*}{ECL}} & Params(M) & 7.94 & 2.75 & 1.39 & 9.68 & 4.83 & 12.12 & 0.12 & 0.02 & 300.58 & 3.76 & 0.42 \\
\multicolumn{1}{c|}{} & FLOPs(G) & 3.87 & 0.41 & 5.07 & 13.54 & 1.87 & 5.55 & 1.74 & 0.01 & 293.98 & 12.48 & 0.13 \\
\multicolumn{1}{c|}{} & T.T.(ms/iter) & 41.67 & 51.96 & 78.25 & 587.51 & 10.66 & 26.59 & 57.09 & 1.37 & 509.17 & 65.83 & 3.78 \\
\multicolumn{1}{c|}{} & T.M.(GB) & 1.12 & 1.13 & 5.08 & 13.15 & 0.70 & 1.24 & 4.24 & 0.04 & 5.82 & 3.02 & 0.30 \\
\multicolumn{1}{c|}{} & I.T.(ms/iter) & 14.21 & 25.66 & 28.21 & 192.42 & 3.46 & 7.08 & 22.79 & 0.20 & 157.13 & 20.52 & 0.48 \\
\multicolumn{1}{c|}{} & I.M.(GB) & 0.47 & 0.47 & 0.63 & 0.94 & 0.23 & 0.26 & 0.82 & 0.02 & 1.41 & 0.93 & 0.23 \\ \midrule
\multicolumn{1}{c|}{\multirow{6}{*}{Exchange}} & Params(M) & 1.51 & 0.67 & 1.39 & 8.40 & 4.83 & 8.59 & 0.08 & 0.02 & 299.94 & 3.76 & 0.42 \\
\multicolumn{1}{c|}{} & FLOPs(M) & 23.20 & 9.29 & 114.83 & 380.79 & 38.88 & 135.01 & 12.25 & 0.15 & 287,909.00 & 311.08 & 3.36 \\
\multicolumn{1}{c|}{} & T.T.(ms/iter) & 5.43 & 27.83 & 35.57 & 87.58 & 10.14 & 28.56 & 27.46 & 1.47 & 505.39 & 9.75 & 3.51 \\
\multicolumn{1}{c|}{} & T.M.(GB) & 0.12 & 0.05 & 0.17 & 0.47 & 0.21 & 0.24 & 0.06 & 0.02 & 5.80 & 0.15 & 0.03 \\
\multicolumn{1}{c|}{} & I.T.(ms/iter) & 5.43 & 6.00 & 6.02 & 24.62 & 1.70 & 6.86 & 3.81 & 0.21 & 156.33 & 2.47 & 0.48 \\
\multicolumn{1}{c|}{} & I.M.(GB) & 0.06 & 0.02 & 0.98 & 0.07 & 0.16 & 0.08 & 0.02 & 0.01 & 1.40 & 0.05 & 0.02 \\ \midrule
\multicolumn{1}{c|}{\multirow{6}{*}{Traffic}} & Params(M) & 5.57 & 7.72 & 0.98 & 14.65 & 4.83 & 11.09 & 0.12 & 0.02 & 301.69 & 3.76 & 0.42 \\
\multicolumn{1}{c|}{} & FLOPs(G) & 7.63 & 1.36 & 10.47 & 51.52 & 6.45 & 14.78 & 4.66 & 0.02 & 288.72 & 33.52 & 0.36 \\
\multicolumn{1}{c|}{} & T.T.(s/iter) & 0.16 & 0.17 & 0.18 & 1.81 & 0.04 & 0.10 & 0.17 & 0.00 & 5.86 & 0.17 & 0.01 \\
\multicolumn{1}{c|}{} & T.M.(GB) & 3.70 & 3.04 & 9.50 & 10.49 & 2.95 & 5.76 & 11.33 & 0.07 & 5.86 & 8.01 & 0.78 \\
\multicolumn{1}{c|}{} & I.T.(ms/iter) & 54.28 & 80.83 & 61.21 & 151.24 & 13.45 & 31.60 & 67.21 & 0.20 & 15.64 & 55.10 & 3.48 \\
\multicolumn{1}{c|}{} & I.M.(GB) & 2.08 & 1.21 & 1.69 & 0.75 & 1.27 & 1.84 & 2.20 & 0.04 & 1.42 & 2.44 & 0.59 \\ \midrule
\multicolumn{1}{c|}{\multirow{6}{*}{Weather}} & Params(M) & 3.19 & 0.75 & 0.03 & 11.21 & 4.83 & 0.50 & 0.10 & 0.02 & 299.97 & 3.76 & 0.42 \\
\multicolumn{1}{c|}{} & FLOPs(G) & 0.11 & 0.02 & 0.04 & 1.29 & 0.10 & 0.01 & 0.05 & 0.00 & 291.21 & 0.82 & 0.01 \\
\multicolumn{1}{c|}{} & T.T.(ms/iter) & 25.95 & 27.85 & 28.69 & 91.59 & 9.45 & 37.85 & 33.30 & 1.38 & 505.88 & 9.89 & 3.53 \\
\multicolumn{1}{c|}{} & T.M.(GB) & 0.18 & 0.09 & 0.05 & 1.21 & 0.22 & 0.05 & 0.16 & 0.02 & 5.80 & 0.02 & 0.04 \\
\multicolumn{1}{c|}{} & I.T.(ms/iter) & 4.38 & 6.00 & 4.11 & 29.73 & 1.66 & 7.19 & 5.10 & 0.20 & 156.27 & 2.96 & 0.48 \\
\multicolumn{1}{c|}{} & I.M.(MB) & 79.81 & 39.97 & 14.45 & 125.18 & 34.88 & 18.28 & 36.03 & 9.06 & 1,397.18 & 90.78 & 24.37 \\ \bottomrule
\end{tabular}
\label{Efficiency_app}
\end{table*}

\section{Model Efficiency}\label{Model_Efficiency_app}
Table \ref{Efficiency_app} presents a comprehensive computational profiling of FAiT against baseline models, encompassing critical efficiency metrics such as parameter count, FLOPs, training/inference time per epoch (T.T./I.T.), and memory consumption (T.M./I.M.). Empirical results demonstrate that FAiT strikes an optimal equilibrium between forecasting accuracy and computational efficiency. This efficiency is primarily attributed to our streamlined architectural design: the Inverted Attention mechanism is built directly upon the vanilla attention module, introducing no additional parameters or complexity, while the DTFM strategy efficiently aggregates and refines spectral information using a lightweight network without incurring excessive overhead. Consequently, FAiT exhibits superior scalability compared to other Transformer-based predictors, a distinct advantage that becomes particularly pronounced on large-scale, high-dimensional multivariate datasets such as ECL and Traffic.


\end{document}